\PassOptionsToPackage{dvipsnames,svgnames,table,xcdraw}{xcolor}
\documentclass[preprint,12pt]{elsarticle}

\usepackage{amssymb}
\usepackage{amsmath}

\usepackage{hyperref}       
\usepackage[final]{changes} 
\usepackage{url}            
\usepackage{wrapfig}
\usepackage{booktabs}       
\usepackage{amsfonts,bm}       
\usepackage{nicefrac}       
\usepackage{microtype}      
\usepackage{xcolor}
\usepackage{ulem}
\usepackage[inline, shortlabels]{enumitem}
\usepackage{todonotes}
\usepackage{changes}
\usepackage{float,subcaption}
\usepackage{caption}
\usepackage{todonotes}
\usepackage{makecell}
\usepackage{tabulary}
\usepackage{tabularx}
\usepackage{booktabs}
\usepackage{graphicx,scalerel}
\usepackage{stackengine}
\usepackage{adjustbox}
\usepackage{diagbox}
\usepackage{tabstackengine}
\usepackage{amsmath,bm}
\usepackage{multirow}
\usepackage{footnote}
\usepackage{array}
\usepackage{ragged2e}
\usepackage{dsfont}
\usepackage{listings} 
\usepackage{soul}

\usepackage{eurosym}
\usepackage{tikz}
\usetikzlibrary{tikzmark}

\def\BibTeX{{\rm B\kern-.05em{\sc i\kern-.025em b}\kern-.08em
    T\kern-.1667em\lower.7ex\hbox{E}\kern-.125emX}}

\newcommand*{\myfullcircle}[1]{
    \begin{tikzpicture}[scale=0.1, line width=.3mm]%
    \draw[black, fill=#1] (0,0) circle (1.5);
    \end{tikzpicture}
}

\colorlet{punct}{red!60!black}
\definecolor{background}{HTML}{EEEEEE}
\definecolor{delim}{RGB}{20,105,176}
\colorlet{numb}{magenta!60!black}

\lstdefinelanguage{json}{
    basicstyle=\tiny\normalfont\ttfamily,
    numbers=left,
    numberstyle=\tiny,
    stepnumber=1,
    numbersep=8pt,
    showstringspaces=false,
    breaklines=true,
    tabsize=3,
    frame=lines,
    backgroundcolor=\color{background},
    literate=
     *{0}{{{\color{numb}0}}}{1}
      {1}{{{\color{numb}1}}}{1}
      {2}{{{\color{numb}2}}}{1}
      {3}{{{\color{numb}3}}}{1}
      {4}{{{\color{numb}4}}}{1}
      {5}{{{\color{numb}5}}}{1}
      {6}{{{\color{numb}6}}}{1}
      {7}{{{\color{numb}7}}}{1}
      {8}{{{\color{numb}8}}}{1}
      {9}{{{\color{numb}9}}}{1}
      {:}{{{\color{punct}{:}}}}{1}
      {,}{{{\color{punct}{,}}}}{1}
      {\{}{{{\color{delim}{\{}}}}{1}
      {\}}{{{\color{delim}{\}}}}}{1}
      {[}{{{\color{delim}{[}}}}{1}
      {]}{{{\color{delim}{]}}}}{1},
}

\DeclareFixedFont{\ttb}{T1}{txtt}{bx}{n}{7} 
\DeclareFixedFont{\ttm}{T1}{txtt}{m}{n}{7}  

\usepackage{color}
\definecolor{deepblue}{rgb}{0,0,0.5}
\definecolor{deepred}{rgb}{0.6,0,0}
\definecolor{deepgreen}{rgb}{0,0.5,0}

\newcommand\pythonstyle{\lstset{
language=Python,
basicstyle=\ttm\linespread{4.2},
morekeywords={self},              
keywordstyle=\ttb\color{deepblue},
emph={MyClass,__init__},          
emphstyle=\ttb\color{deepred},    
stringstyle=\color{deepgreen},
frame=lines,
showstringspaces=false,
breaklines=true,
numbers=left,
numberstyle=\tiny,
stepnumber=1,
numbersep=8pt,
tabsize=1,
backgroundcolor=\color{background},
commentstyle=\tiny\color[HTML]{228B22}\sffamily,
literate={\ \ }{{\ }}1
}}

\lstnewenvironment{python}[1][]
{
\pythonstyle
\lstset{#1}
}
{}

\newcommand\pythoninline[1]{{\pythonstyle\lstinline!#1!}}

\journal{Energy and AI}

\begin{document}

\begin{frontmatter}

\title{Machine Learning for Physical Simulation Challenge Results and Retrospective Analysis: Power Grid Use Case}

\author[C,IRT]{Milad Leyli-abadi} 
\author[RTE]{Jérôme Picault}
\author[RTE]{Antoine Marot}
\author[IRT]{Jean-Patrick Brunet}
\author[IRT]{Agathe Gilain}
\author[ASU]{Amarsagar Reddy Ramapuram Matavalam}
\author[ASU]{Shaban Ghias Satti}
\author[XJTU]{Qingbin Jiang}
\author[XJTU]{Yang Liu}
\author[UT]{Dean Justin Ninalga}

\affiliation[C]{organization={Corresponding author},
            email={milad.leyli-abadi@irt-systemx.fr}}
            
\affiliation[IRT]{organization={IRT SystemX},
            city={Palaiseau},
            state={Ile-de-France},
            country={France}}

\affiliation[RTE]{organization={Réseau de Transport d'Electricité (RTE)},
            city={Paris La Défense},
            state={Ile-de-France},
            country={France}}

\affiliation[ASU]{organization={Arizona State University},
                  city={Tempe},
                  state={Arizona},
                  country={USA}}
                  
\affiliation[XJTU]{organization={Xi'an Jiaotong University},
                  addressline={MOE KLINNS Lab, Faculty of Electronic and Information Engineering}, 
                  city={Xi’an},
                  state={Shaanxi},
                  country={China}}

\affiliation[UT]{organization={University of Toronto},
                  state={Toronto},
                  country={Canada}}

\begin{abstract}
This paper addresses the growing computational challenges of power grid simulations, particularly with the increasing integration of renewable energy sources like wind and solar. As grid operators must analyze significantly more scenarios in near real-time to prevent failures and ensure stability, traditional physical-based simulations become computationally impractical. To tackle this, a competition was organized to develop AI-driven methods that accelerate power flow simulations by at least an order of magnitude while maintaining operational reliability. This competition utilized a regional-scale grid model with a 30\% renewable energy mix, mirroring the anticipated near-future composition of the French power grid. A key contribution of this work is through the use of LIPS (Learning Industrial Physical Systems), a benchmarking framework that evaluates solutions based on four critical dimensions: machine learning performance, physical compliance, industrial readiness, and generalization to out-of-distribution scenarios. The paper provides a comprehensive overview of the Machine Learning for Physical Simulation (ML4PhySim) competition, detailing the benchmark suite, analyzing top-performing solutions that outperformed traditional simulation methods, and sharing key organizational insights and best practices for running large-scale AI competitions. Given the promising results achieved, the study aims to inspire further research into more efficient, scalable, and sustainable power network simulation methodologies.
\end{abstract}


\end{frontmatter}




\section{Introduction}
Physical simulations are a fundamental tool for analyzing and optimizing complex industrial systems, such as power grid management, aeronautics, gas production, and thermal comfort. These simulations play a critical role throughout the system life-cycle, from initial design and verification to real-time decision-making. However, traditional numerical simulation methods, while accurate, often suffer from high computational costs, making them prohibitive for large-scale, real-time applications. This limitation is particularly evident in power grid operations, where Transmission System Operators (TSOs) must continuously assess grid stability, anticipate potential failures, and implement remedial actions. Simulating the grid state requires solving non-linear, non-convex power flow equations \cite{molzahn2019survey}, which is computationally intensive. 

To address the computational challenges of physical simulations, there is a growing interest in using machine learning techniques \cite{carleo2019machine}. These techniques aim to accelerate computation while maintaining acceptable accuracy. Deep Neural Networks (DNNs) have shown promise in various domains \cite{tompson2016, kasim2021building, donnot2019leap, lam2023learning}, offering significant speed-ups by replacing some computational components with data-driven numerical models. These approaches often involve emulating existing simulators through supervised learning or developing new types of differential solvers in an unsupervised manner, sometimes falling under the category of Physics-Informed Machine Learning \cite{raissi2017physics}. These hybrid approaches seek to combine the strengths of traditional physical models with the efficiency of machine learning which are also known as Scientific Machine Learning (SciML).

The integration of machine learning in physics-based simulations has gained traction and several challenges have been organized to benchmark and improve hybrid modeling techniques. For instance, previous editions of the Machine Learning for Physical Simulation (ML4PhySim) challenge have explored applications in aerodynamics and computational fluid dynamics (CFD) \cite{yagoubi2024neurips, yagoubi2024ml4physim}. In the power grid sector, various competitions have been organized which addressed the control problems in power grid domain \cite{marot2020learning, marot2021learning, marot2022learning}. It was required to develop a Reinforcement Learning (RL) agent to maintain the grid stability under various configurations. However, these efforts have primarily focused on control and optimization rather than surrogate modeling for power flow simulation.

This challenge is significant as it represents the first dedicated effort to explore hybrid modeling techniques for power grid simulations. Using the experience gained during the organization of the previous ML4PhySim challenge, a similar evaluation and infrastructure pipeline is used for the organization of this challenge. However, this challenge considered a completely different industrial domain with specific underlying physics. The expert knowledge of TSOs was valuable for the definition of the use case scenario and the adjustment of domain-specific evaluation criteria. 

The considered use case was related to the growing emergence of renewable energy sources that has introduced unprecedented levels of variability and uncertainty into power systems, creating a need for more frequent and computationally efficient simulations \cite{marot2022perspectives}. TSOs face the critical task of maintaining grid stability and resilience during the transition to low-carbon energy. While efficient power flow simulations are vital for grid reliability, energy distribution optimization, and failure prevention, current physical solvers hardly meet the escalating demand for swift and accurate grid assessments in real-time operational contexts \cite{marot2020towards}. In this context, an acceleration by several order of magnitudes is expected.

To address these challenges and inspire academic and industry researchers specializing in AI, this competition aimed to develop innovative machine learning (ML) surrogate models for computing grid power flows. The goal was to optimize the trade-off between inference accuracy and computational efficiency. The competition’s scope can be summarized as follows:
\begin{itemize}
    \item First competition to motivate the use of hybrid models for power grid systems;
    \item Promote reproducibility, transparency, and industrial relevance by incorporating a dedicated challenge phase focused on end-to-end reproducibility of machine learning for physics. Participants were therefore required to submit untrained models, which were then fully retrained and evaluated on the organizers' servers. This ensured that the models were reproducible and could be tested independently, thereby enhancing the transparency and fairness of the entire process;   
    \item Encourage finding synergies in terms of solutions for different physical problems, in particular in relationship with the previous edition of the ML4PhySim challenge focused on the Airfoil use case;
    \item Encourage collaboration between AI and physical sciences by designing new ML approaches (algorithms, architectures) targeting physical problems;
    \item Ensure a meaningful and standardized evaluation of the submitted surrogate models through a comprehensive set of evaluation criteria proposed in our testbed framework;
    \item Benchmark all submissions on the same environment and infrastructure, especially for fair speed-up comparisons.
\end{itemize}

Building on the experience gained from organizing both previous challenges and this one in particular, we propose a comprehensive analysis of the outcomes from multiple perspectives. In addition to a technical evaluation of the winning solutions, we examine the effectiveness of the competition materials provided to participants, the robustness of the evaluation methodology used to generate the leaderboard rankings, and the incentive strategies implemented to foster broader participation. This analysis offers valuable insights into the key success factors of the challenge and highlights areas for potential improvement in future editions.

The rest of this paper is structured as follows: Section \ref{sec:overview} provides an overview of the competition, detailing the problem statement, objectives, design process, configurations, datasets, evaluation pipeline, and available resources. Section \ref{sec:outcomes_participants} presents the competition outcomes, including the final leaderboard and an in-depth description of the winning solutions, along with a deeper analysis of the results. Section \ref{sec:outcomes_organization} discusses the organizational outcomes, sharing insights and lessons learned from the competition. Finally, Section \ref{sec:conclusions} concludes the paper.

\section{Competition overview}
\label{sec:overview}
This section describes the different steps of the competition design, including the problem overview, the task studied, the evaluation testbed, the competition resources and execution pipeline and the extra materials that were made available to the participants. 

\subsection{Problem overview}
Several problems may occur in a power grid. The role of a TSO (Transmission System Operator) is, among others, to ensure at any time the safety of the power grid. Operators have to face unexpected events (losing a line for example due to weather constraints) or to anticipate events such as variation of production during the day or as equipment’s maintenance. They do so by assessing the risks, leveraging grid flexibility through simulations and carefully choosing sets of remedial actions which act on the grid topology or on the production levels.

This competition focused on this particular aspect of risk assessment in power grids. The problem is to be robust against the unintentional contingencies that may occur on the power grid (several hours ahead to near real-time) and warn the operators accordingly to cover the associated risk. To do so, current physical simulators simulate incidents (aka contingencies) involving various elements of the grid (such as the disconnection of a line/production unit), one by one. For each contingency, a risk (weakness of the grid) is identified, when overloads are detected by the simulation engine on some lines. On a real grid, this scenario means running millions of daily simulations for a 24 hour forecasted time horizon. This simulation becomes even more costly with traditional power flow methods as the grid topology changes over the day resulting in evolving problem size/dimensions. Thus, computation time is critical, especially since this risk assessment is refreshed every 15 minutes. In this context, the computational cost of physical solvers becomes a bottleneck, restricting the necessary exploration of potential solutions.    

\subsection{Physical power flow simulators}
In power grid operation, the risk assessment is done through power flow computation using a physical simulator. At its core, this simulator takes as \textit{input variables} the amount of \textit{productions} and \textit{loads} in the grid as well as the \textit{topology} vector and computes some necessary physical variables of the grid state that are necessary to assess the security of the grid, i.e., \textit{voltages}, \textit{angles}, \textit{active and reactive power}, \textit{intensity} at each node of the grid (see Table \ref{tbl: notation} for notations). The computation of the grid state involves a set of physical laws such as:
\begin{itemize}
    \item Kirchhoff's current law (states that the current flowing into a node must be equal to the current flowing out of it);
    \item Global energy conservation law (states that the sum of all productions must be equal to the sum of all loads and losses);
    \item Joule effect  (states that the power of heating generated by an electrical conductor is proportional to the product of its resistance and the square of the current);
    \item Physical constraints related to generations with their upper and lower limits.
\end{itemize}

More specifically, the physical resolution of the problem is derived from a set of powerflow equations \cite{molzahn2019survey} described at any node $k$ of the grid. The power injected at a node of the network $s_k$ is the sum of active ($p_k$) and reactive powers ($q_k$): $s_k = p_k + q_k$. From Kirchhoff energy conservation law, the relation between voltage angle and magnitude can be formulated for node $k$ and neighboring nodes $m$ as follows:

\hspace{-0.6cm}
\resizebox{.82\linewidth}{!}{
\begin{minipage}{\linewidth}
\begin{eqnarray}
\label{eq:Power_equations}
\left\{
\begin{array}{lll}
0 = -&{p_k} + \sum_{m=1}^K {|v_k|} {|v_m|} (g_{k,m}\cdot \cos ({\theta_k} - {\theta_m}) + b_{k,m} \sin ({\theta_k} - {\theta_m})) & \text{Active power;}\vspace{0.2cm} \\
0 = &{q_k} + \sum_{m=1}^K {|v_k|} {|v_m|} (g_{k,s}\cdot \sin ({\theta_k} - {\theta_m}) - b_{k,m} \cos ({\theta_k} - {\theta_m})) & \text{Reactive power,}
\end{array}
\right.
\end{eqnarray}
\end{minipage}
}\\

\noindent where phasors $\theta_k$ are unknown for all node $k$; either voltages $|v_k|$ or reactive powers ${q_k}$ are known inputs at any given node $k$; active powers {$p_k$} are known inputs except for slack generator which can compensate for estimated energy losses, however, in our dataset we indeed use the re-estimated production inputs rather than the raw inputs; $g_{k,m}$,  $b_{k,m}$ are known line characteristics for all nodes. For each line $\ell$, the active $p^\ell$ and the reactive $q^\ell$ power flows or the current $a^\ell$ can be derived using the Ohm's law. This problem is non-linear and non-convex. To estimate these variables, a power flow solver such as LightSim2grid \cite{lightsim2grid} can be used. The physics solvers are generally based on iterative optimization algorithms such as Newton-Raphson. Further details on the power flow equations could also be found in \ref{app2}.

\paragraph{\textbf{Limitations of physical simulators and motivations}} 
Currently, security analysis\footnote{An essential tool for assessing the security of power systems, aimed at determining whether and to what extent a power system remains resilient to disturbances.} relies on physical simulators, which are primarily limited to N-1 contingency cases (ensuring grid stability in the event of a single power line disconnection). These simulators often run brute-force simulations, even for scenarios that are clearly safe. Machine learning could enhance this process by enabling fast contingency screening \cite{donnot2018anticipating}, allowing for the early identification of potentially risky cases and reducing the number of unnecessary simulations. This approach could be extended to analyze both N-1 and N-2 contingencies, as well as scenarios involving power grid topology variations.

\paragraph{\textbf{Competition Objective}} The objective in this competition was to accelerate costly physical simulations using AI-augmented simulators, while considering the compliance with physical laws.

\subsection{Competition design and datasets}
This section describes the power grid environment that is considered for this competition along with the specific configurations, scenarios, and datasets provided to the participants.

\begin{figure}
    \centering
    \includegraphics[width=1\linewidth]{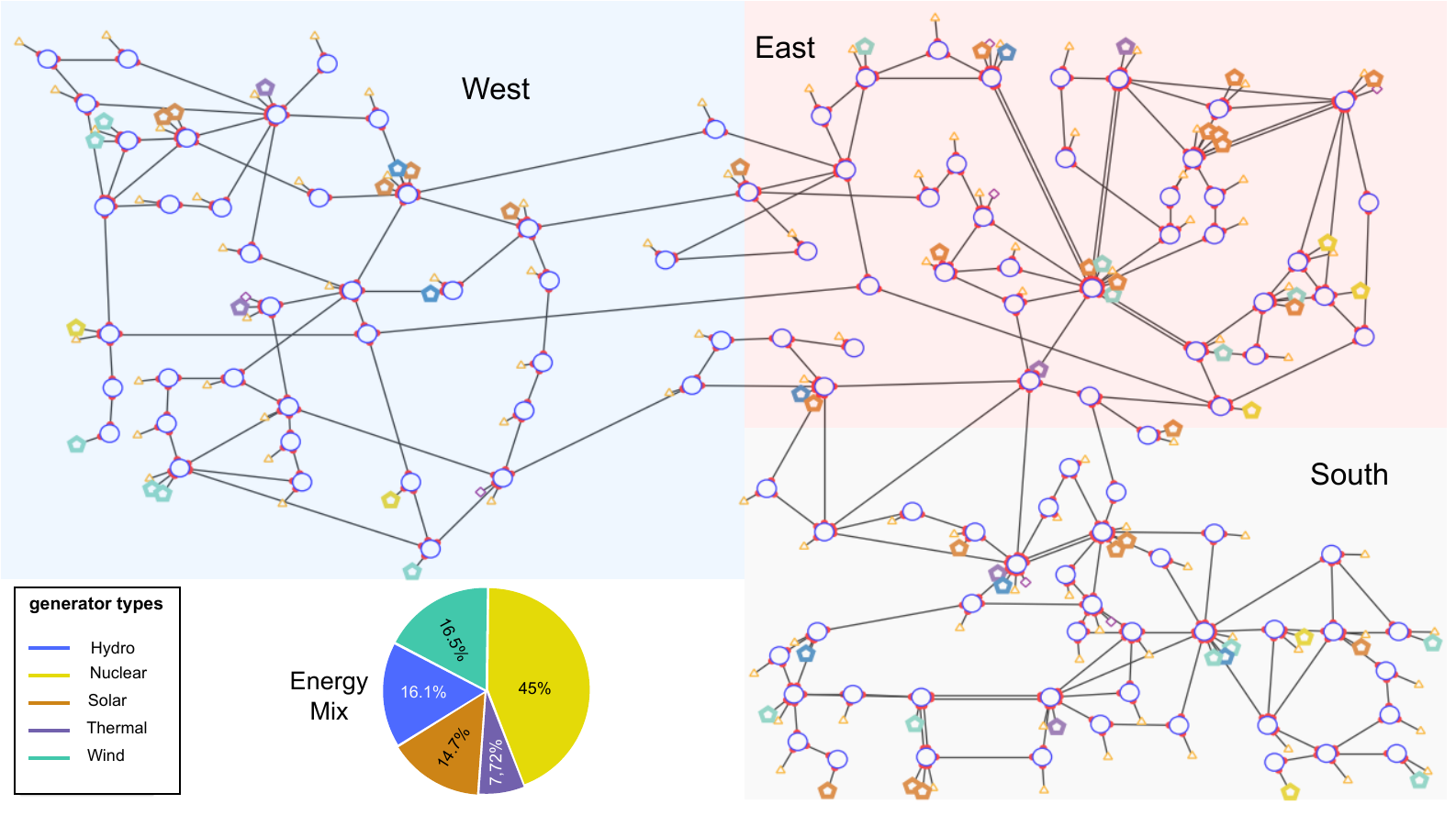}
    \caption{Power grid environment used for the competition (\textsc{L2RPN\_IDF\_2023}).}
    \label{fig:env_idf}
\end{figure}

\paragraph{\textbf{Environment and configurations}}
In this challenge, we opted for a recent power grid environment which is based on IEEE 118 grid with high renewable penetration \cite{pena2017extended}. This grid along the productions and loads is shown in Figure \ref{fig:env_idf}. It includes 118 substations connected through 186 power lines. There are also different regions on this grid, with their specific distribution of productions and loads, which makes the power flow computation more challenging. The overall distribution of energy mix is also summarized using the pie chart in this figure. As can be seen, a high proportion of productions are based on renewable energy sources. The environment and related configurations are imported using the Grid2Op Python package \cite{grid2op}. 

Toward real-world applications near real-time, we include the following considerations in the competition setup: 
\begin{itemize}
    \item Penetration of renewable energy corresponding to 30\% wind and solar energies (see Figure \ref{fig:env_idf});
    \item Changing topologies at substations as a key operational consideration for flexible grids. As can be seen in Figure \ref{fig:topology_reconfiguration}, the topological changes (allowing to assign a a specific busbar to each element of the grid connected to the substations) may be used as remedial actions to avoid the power lines overload problem. Hence, we include such topology reconfigurations in our scenarios, as the powerflow simulators (augmented by AI or not) should be robust against them;
    \begin{figure}
        \centering
        \includegraphics[width=0.95\linewidth]{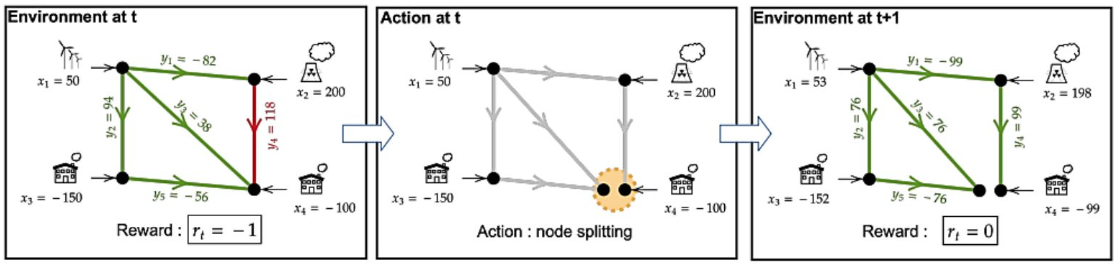}
        \caption{Illustration of a remedial action through a topological change at a substation. A topological action splitting a substation at time stamp $t$ allows to resolve the overload problem on one of the power lines.}
        \label{fig:topology_reconfiguration}
    \end{figure}
    \item Trust from the operators with acceptable compliance to physical laws.
\end{itemize} 

\paragraph{\textbf{Datasets and scnenarios}}
\label{par:datasets}
For all the datasets, a \textit{reference} configuration is considered as the starting point and on which a set of variations are applied (here some topology changes at different substations of the power grid). This corresponds to the \textit{standard} variability that human operators are facing in their daily operations (the goal was to have realistic starting points). More specifically, we authorized 4 different topology changes acting on substations 1, 16 and 28 by limiting the maximum allowed simultaneous changes to 2 at each sample without power line disconnections. Hence, each dataset could randomly integrate or not these topology changes during the data generation process with 10\% of probability to not include any reference topology changes. In \ref{app:generation_example}, we provide an example on how to design and generate datasets. More details concerning the datasets and their scenarios used for learning and evaluation of models are provided in the following. 

\begin{itemize}
    \item Training dataset: An ensemble of 300\,000 samples is generated for the training purpose of the models. Based on our experimentation, this amount of data provides an optimal balance of grid configuration diversity needed for learning. They are generated from a specific subset of chronics\footnote{In the Grid2Op context, the data is organized as time series describing the electricity injections into the powergrid, referred to as chronics.} and with specific seeds to ensure reproducibility. In addition to the above-mentioned reference topology, we allow only one power line disconnection with the probability of $0.7$. Hence, the training dataset comprises whether samples with only reference topology or samples including one power line disconnection in addition to reference topology. This also follows the goal of the security analysis discussed previously.
    
    \item Validation and test dataset: An ensemble of 100\,000 samples are generated for each of the validation and test datasets. As for the training dataset, specific chronics and seeds are used for each of these datasets. The validation and test datasets share the same topology distribution. In addition to the reference topology, a power line is randomly disconnected for each sample. A sample of this dataset (with a focus on the disconnection region) is shown in Figure \ref{fig:test_data_scenario}.
    \begin{figure}[H]
        \centering
        \includegraphics[width=0.65\linewidth]{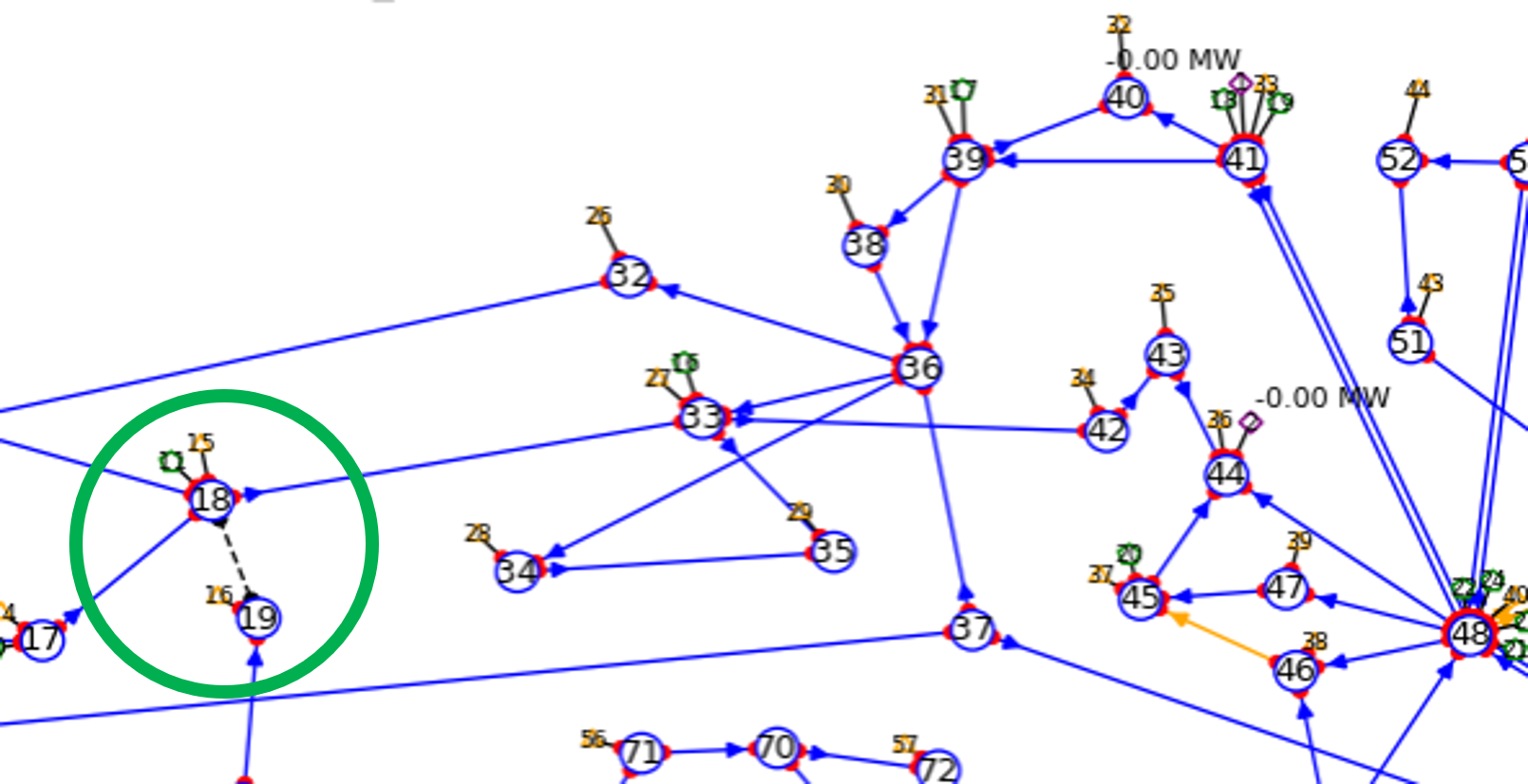}
        \caption{Test dataset distribution. Only one disconnected line is authorized for validation and test datasets, as is also the case for the training dataset.}
        \label{fig:test_data_scenario}
    \end{figure}
    
    \item Out-Of-Distribution (OOD) Test dataset: An ensemble of 200\,000 samples are generated for OOD test dataset. As for the previous datasets, we considered a specific chronic range and seeds. In order to have a slightly different distribution from the training and test datasets, and to assess the generalization capability of proposed solutions in the competition, \textit{two} simultaneous power lines are disconnected this time. Furthermore, in order to increase the non-linear impact of these disconnections, they are implemented in the same region of the power grid. A sample of this dataset is shown in Figure \ref{fig:test_ood_scenario}. We can see in this figure, that the disconnected power lines are close to each other. 
    \begin{figure}[H]
        \centering
        \includegraphics[width=0.65\linewidth]{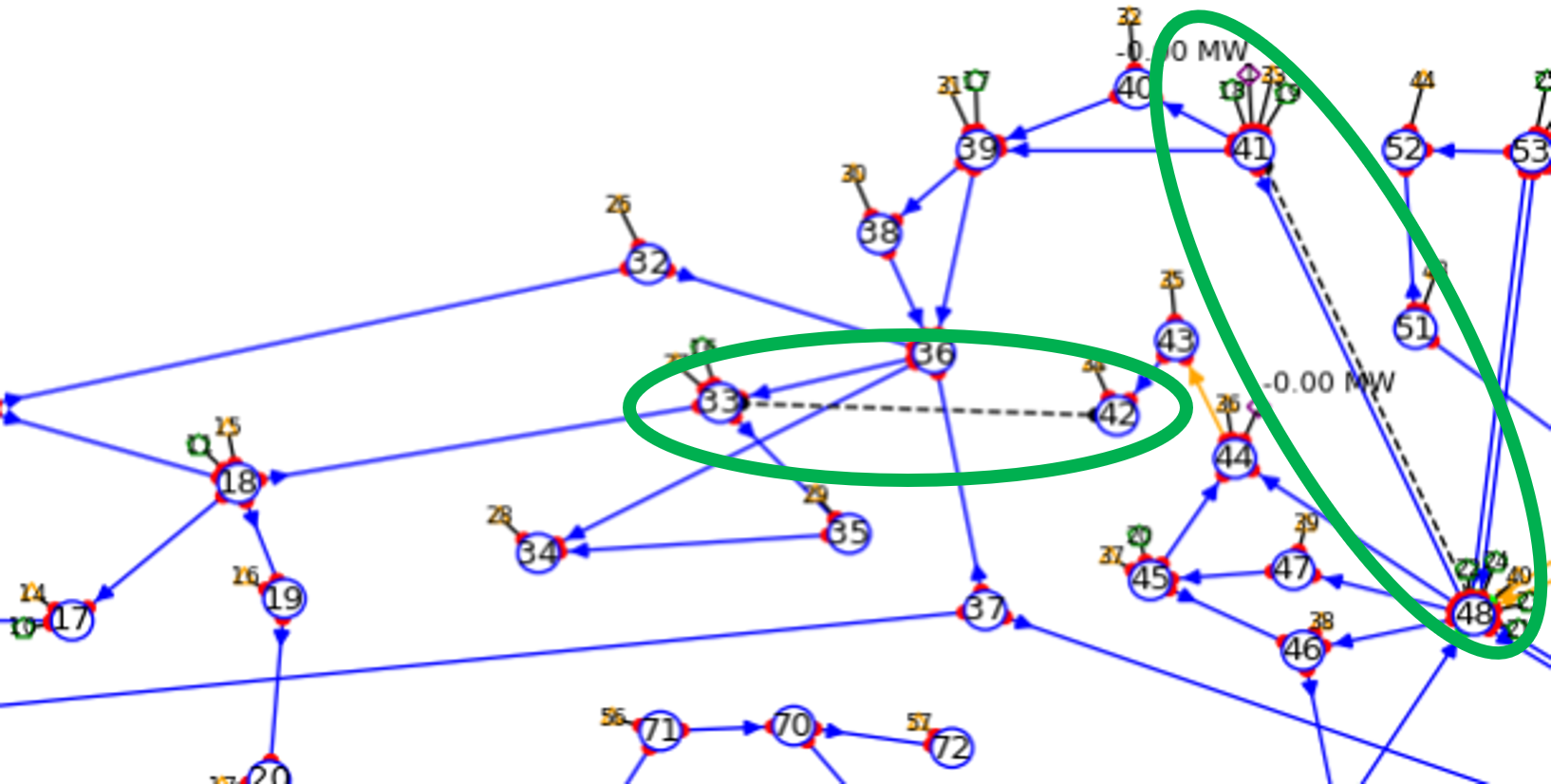}
        \caption{Out-of-Distribution test dataset distribution. Two powerlines are disconnected simultaneously, which is a new topology that was never observed during the training.}
        \label{fig:test_ood_scenario}
    \end{figure}
\end{itemize}

\paragraph{\textbf{Inputs and outputs}}
\label{par:inputs_outputs}
In this competition, the objective was to predict the powerflows from the injections at different substations. For this purpose, some information concerning the topology, connectivity of the powerlines along with some physics-based information were made available to the participants. These could be leveraged for more advanced Physics Informed Machine Learning solutions. These variables are summarized in Table \ref{tab:inputs_outptus}. 

Two main possible configurations in terms of inputs/outputs for the surrogate solutions were:
\begin{itemize}
    \item Develop an ML-based approach that takes the injections as inputs and outputs directly the required power flow variables. The topology information could also be considered among the inputs or exploited to develop for example graph-based approaches;
    \item Develop an physics informed surrogate model approach which takes the injection as inputs and leverage physics attributes such admittance matrix and outputs the voltage angles and magnitude. From these quantities the power flow could be derived easily using different physical laws formulas.
\end{itemize}

\begin{table}[H]
\caption{Input and outputs of the problem. The topology and physics attributes could also be exploited as inputs.}
\resizebox{\linewidth}{!}{
\begin{tabular}{l|l}
\toprule
\multicolumn{1}{c|}{\textbf{Variable}} & \multicolumn{1}{c}{\textbf{Description}} \\
\midrule
\midrule
\multicolumn{2}{c}{\textit{Inputs}} \\                       
\hline
$prod_p$ &  Generator active power \\
$prod_v$ &  Generator voltage \\
$load_p$ &  Load active power \\      
$load_q$ &  Load reactive power \\
\midrule
\midrule
\multicolumn{2}{c}{\textit{outputs}} \\ \hline
$a_{or}$ & Currents at the origin of the power lines \\
$a_{ex}$ & Currents at the extremity of the power lines \\
$p_{or}$ & Active power at the origin of the power lines \\
$p_{ex}$ & Active power at the extremity of the power line \\
$v_{or}$ & Voltage at the origin of the power line \\
$v_{ex}$ & Voltage at the extremity of the power line\\
\midrule
\midrule
\multicolumn{2}{c}{\textit{Topology}} \\ \hline
Line status & A vector including the status of the lines (connected/disconnected)\\
Topology vector & A vector indicating the bus connectivity of each element\\ 
\midrule
\midrule
\multicolumn{2}{c}{\textit{Physics attributes}} \\ \hline
$YBus$ & The admittance matrix including the admittances between each pair of nodes\\
$SBus$ & The power matrix including the power values between each pair of nodes\\
& The indices of P-V nodes. A P-V node is a bus where active power (P) and voltage (V) \\
$P-V nodes$ & are specified, while reactive power (Q) adjusts to maintain voltage stability, typically \\
& associated with generators.\\
$Slack$ & The indices of slack nodes\\
\bottomrule
\end{tabular}}
\label{tab:inputs_outptus}
\end{table}

\subsection{Evaluation, ranking and competition resources}
This section describes the different competition stages, the evaluation pipeline along with the set of KPIs that are used to assess the performance of submissions, the scoring process, the infrastructure, and the extra materials which were provided to the participants facilitating the comprehension of the use case and submission procedure.

\paragraph{\textbf{Competition phases}}
The competition began on April 16, 2024, and concluded on September 29, 2024. It was run on Codabench\footnote{\url{https://www.codabench.org/competitions/2378/}} and structured into three distinct phases:  

\begin{itemize}
    \item \textit{Warm-up phase (April 16 – June 24):} Participants familiarized themselves with the provided materials and the competition platform. They made their initial submissions and provided feedback to the organizers. Based on this feedback, we refined the scoring computation, updated certain evaluation aspects of the LIPS framework, and added additional packages to the competition container.  

    \item \textit{Development phase (June 24 – September 25):} This was the main phase of the competition, during which participants developed their solutions and tested their trained models against the provided validation dataset. They had access to the global score of all submissions but not to the internal details of other participants' models. Initially scheduled to conclude at the end of August, this phase was extended until September 25 to allow more time for refinement.  

    \item \textit{Final phase (September 25 – October 10):} This phase was managed solely by the organizers. All submissions were gathered, manually verified, and executed 10 times. The final rankings were determined using the mean and standard deviation of the obtained results. The official results were announced during the AI-day for grid operations\footnote{\url{https://www.tinyurl.com/aiday4powergrid}} organized by RTE.   
\end{itemize}

\begin{figure}[H]
    \centering
    \includegraphics[width=\linewidth]{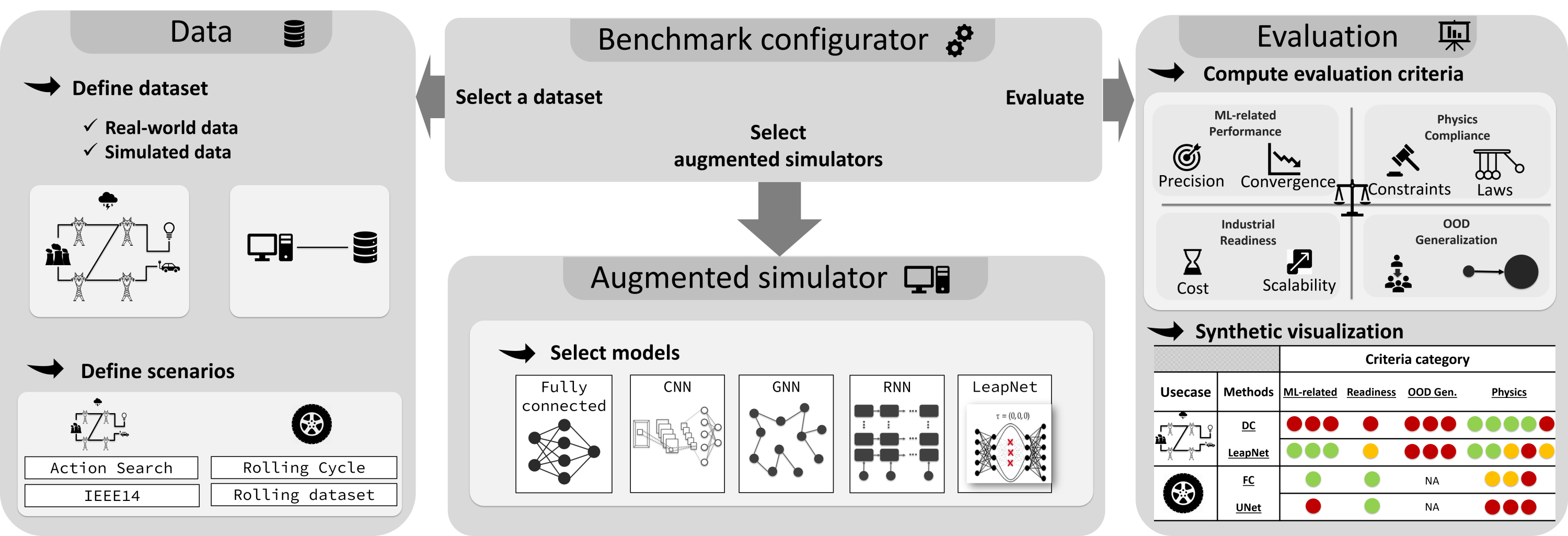}
    \caption{Learning Industrial Physical Simulation (LIPS) framework.}
    \label{fig:lips}
\end{figure}

\paragraph{\textbf{Competition Testbed framework}}
\label{par:lips}
The Learning Industrial Physical Simulation (LIPS) Framework \cite{leyli2022lips} was used as the evaluation pipeline and also for scoring of the submissions. LIPS is a unified and extensible open-source platform\footnote{\url{https://github.com/IRT-SystemX/LIPS}} designed for benchmarking ML-based physical simulations in a uniform yet adaptable manner. It facilitates the evaluation of various ML-based physical simulators through four key modules: data management, benchmark configurator, augmented simulator, and evaluation (Figure \ref{fig:lips}).

Primarily, the LIPS framework is utilized to establish generic and comprehensive evaluation criteria for ranking submitted solutions, beyond the usual ML-related performance. This evaluation process encompasses multiple aspects to address industrial requirements and expectations. The following categories outline the evaluation criteria considered:
\begin{itemize}
\item{ML-related performance}: we focus on assessing the accuracy of augmented simulators using classical ML metrics. The Mean Absolute Percentage Error computed on 10\% highest percentile (MAPE90)\footnote{This variation of MAPE is used as very low currents contribute little to the overall power flow and voltage profile, meaning they do not significantly influence grid stability or efficiency.} is considered for currents ($a_{or}$, $a_{ex}$); The same criteria for 90\% highest quantile (MAPE10\footnote{This variation of MAPE is used to avoid the MAPE criteria explosion due to zero values.}) is considered for the evaluation of active powers ($p_{or}$, $p_{ex}$); Finally, Mean Absolute Error (MAE) is considered for the evaluation of voltages ($v_{or}$, $v_{ex}$);
\item{Physics compliance}: ensuring adherence to physical laws is crucial when simulation results influence real-world decisions. Depending on the benchmark's criticality, this category of criteria aims to specify the types and quantity of physical laws that must be satisfied. Table \ref{tab:Physical_laws} summarizes the physical laws that we have considered in this competition;
\item{Application-based out-of-distribution (OOD) Generalization}: given the necessity in industrial physical simulation to extrapolate over minor variations in problem configuration, we incorporate OOD evaluation, such as unseen power grid topology variations (see Test OOD dataset description in Section \ref{par:datasets});
\item{Industrial readiness}: When deploying a model in real-world applications, it should consider the real data availability and scale-up to large systems. In this competition, for practical necessity and feasibility reasons, we considered only the application time, which is the computation time of a model when integrated in a specific application.
\end{itemize}

\begin{table}[H]
    \centering
    \caption{Physical laws considered for evaluation of physics score.}
    \renewcommand{\arraystretch}{1.3}
    \resizebox{\columnwidth}{!}{
    \begin{tabular}{cllm{5cm}}
        \toprule
        \textbf{ID} & \textbf{Type} & \textbf{Measure} & \textbf{Description} \\
        \hline
        \multicolumn{4}{c}{\textbf{Basic}} \\
        \hline
        \hline
        P1 & Current positivity & $\frac{1}{L}\sum_\ell^L \mathds{1}_{(\hat{a}^\ell_{or,ex} < 0.)}$ & Proportion of negative current \\
        \hline
        P2 & Voltage positivity & $\frac{1}{L}\sum_\ell^L \mathds{1}_{(\hat{v}^\ell_{or,ex} < 0.)}$ & Proportion of negative voltages \\
        \hline
        P3 & Losses positivity & $\frac{1}{L}\sum_\ell^L \mathds{1}_{(\hat{p}^\ell_{ex} + \hat{p}^\ell_{or} < 0.)}$ & Proportion of negative energy losses \\
        \hline
        P4 & Disconnected Line & $\frac{1}{L_{disc}}\sum_{\ell_{disc}}^{L_{disc}} \mathds{1}_{(\lvert\hat{x}^\ell_{ex}\rvert + \lvert\hat{x}^\ell_{or}\rvert > 0.)}$ & Proportion of non-null $a,p$ or $q$ values \\
        \hline
        P5 & Energy Losses & $\frac{\sum_{\ell=1}^L (\hat{p}^{(\ell)}_{ex} + \hat{p}^{(\ell)}_{or})}{Gen} \in [0.005,0.04]$ & Energy losses range consistency \\
        \hline
        \hline
        \multicolumn{4}{c}{\textbf{Uni-dimension Law}} \\
        \hline
        P6 & Global Conservation & $\text{MAPE}((Prod - Load) -  (\sum_{\ell=1}^L (\hat{p}^{\ell}_{ex} + \hat{p}^{\ell}_{or})))$ & Mean energy losses residual \\
        \hline
        P7 & Local Conservation & $\text{MAPE}((p^{prod}_k - p^{load}_k) - (\sum_{l \in neig(k)} \hat{p}^{\ell}_{k}))$ & Mean active power residual at nodes \\
        \hline
        P8 & Joule law & MAPE($\sum_{\ell=1}^L (\hat{p}^\ell_{ex} + \hat{p}^\ell_{or}) - R \times \frac{\sum_{\ell=1}^L (\hat{a}^\ell_{ex} + \hat{a}^\ell_{or})}{2} $) & Joule law considering predicted powers and currents \\
        \bottomrule
    \end{tabular}}
    \label{tab:Physical_laws}
\end{table}

In addition to the evaluation and scoring features used to rank submissions, LIPS integrates the power grid use case along with the physical solver. This integration enables seamless generation or import of data for specific scenarios and associated datasets. Moreover, all developed approaches were required to follow a standardized augmented simulator template within LIPS, ensuring consistency in both development and evaluation procedures. 

\paragraph{\textbf{Scoring}}
To be able to compare through various models and to be able to rank the participants submissions, one solution is to aggregate the introduced evaluation criteria. This aggregation may also take into account the importance of each subcategories of evaluation criteria, which should be adjusted with respect to the context, studied industrial problem and domain experts knowledge. To this end, we propose the following scoring methodology for ranking of different solutions.  

To compare and benchmark through the introduced evaluation criteria categories, in this competition, two different test datasets were provided as mentioned in the previous section, i.e., a \textit{Test dataset} representing the same distribution as the training dataset (only one disconnected power line), and a \textit{OOD test dataset} representing a slightly different distribution from the training set (with two simultaneous disconnected power lines).

The \textit{ML-related} and \textit{Physical compliance} criteria were computed separately on these datasets. The \textit{speed-up} was computed only on test dataset as the inference time may remains same throughout the different dataset configurations. Hence, the global score is calculated based on a linear combination  of the introduced evaluation criteria categories on these datasets:
\begin{eqnarray}
\label{eq:score}
    \text{Score} = \alpha_{test} \times \text{Score}_{{test}} + \alpha_{ood} \times \text{Score}_{{OOD}} + \alpha_{speed-up} \times \text{Score}_{speed-up},
\end{eqnarray}
\noindent where $\alpha_{test}=30\%$, $\alpha_{ood}=30\%$, and $\alpha_{speed-up}=40\%$ are the coefficients to calibrate the relative importance related to metrics computed on test, OOD test dataset and speed-up. The speed-up is measured against a fast state-of-the-art physical solver baseline \cite{lightsim2grid}. We give a slightly more importance to speed-up score, as the main objective of this challenge is to accelerate physical simulations. Each of the scores computed on the test and OOD datasets are also decomposed to machine learning and physics compliance metrics with corresponding sub-coefficients $\alpha_{ML}=66\%$ and $\alpha_{Physics}=34\%$. It prioritize the more accurate solutions by retaining a sufficient contribution of physics compliance. 

The detailed procedure on how to compute each of these three sub-scores is explained in \ref{app:score_formulation}. Various practical examples showing the score computation for different baseline methods are shown in \ref{app:score_example}.

\paragraph{\textbf{Competition resources and execution pipeline}} 
To evaluate contestant submissions, six similar GPUs—two A6000 and four A40 (48GB VRAM)—were made available, sponsored by NVIDIA and IRT SystemX. These GPUs were distributed across two identical servers, each equipped with two Intel Xeon 5315Y processors (8 cores, 16 threads, 3.2GHz) and 256GB DDR4 RAM.

\begin{figure}[htbp]
    \centering
    \includegraphics[width=0.7\linewidth]{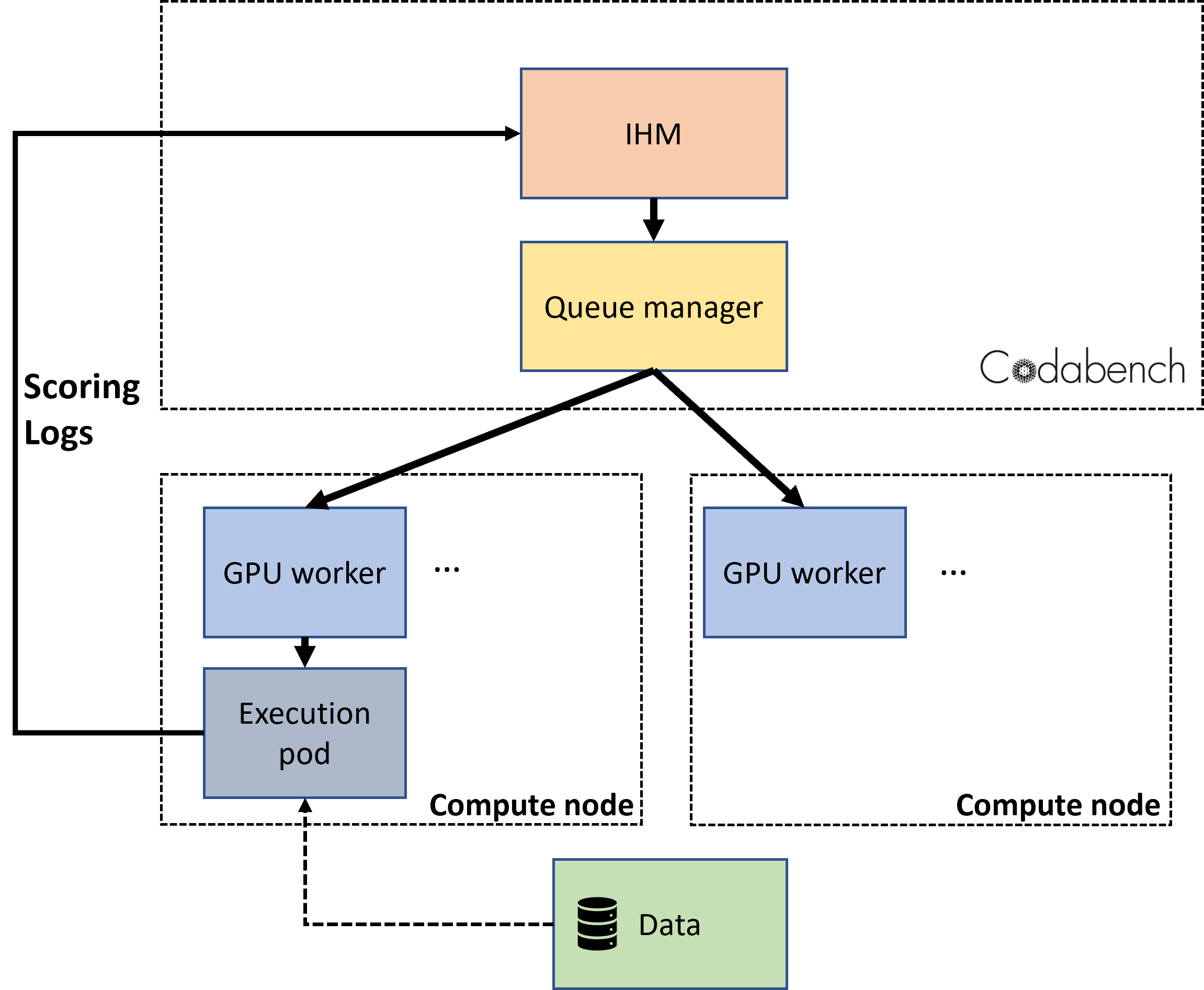}
    \caption{Competition execution pipeline and infrastructure}
    \label{fig:competition_pipeline}
\end{figure}

The competition's execution pipeline and infrastructure, illustrated in Figure \ref{fig:competition_pipeline}, assigned each GPU to a dedicated compute node (worker) with 10 CPU threads. These workers were containerized to ensure complete isolation of each submission during evaluation and scoring. The competition leveraged a submission and execution pipeline originally developed for the ML4CFD competition \cite{yagoubi2024neurips}. Participants submitted their solutions via Codabench\footnote{\url{https://www.codabench.org/competitions/2378/}}, which managed the execution queue, enabling full training, evaluation, and scoring within a standardized environment. Each team was allowed up to 10 submissions per day (with a maximum of 2 simultaneous submissions) and a total of 1,000 submissions per phase.

This setup ensured fair and consistent evaluation, preventing hardware discrepancies from influencing results while allowing participation regardless of individual GPU resources.

\paragraph{\textbf{Competition extra materials}}
A starting kit\footnote{\href{https://github.com/IRT-SystemX/ml4physim\_startingkit\_powergrid}{https://github.com/IRT-SystemX/ml4physim\_startingkit\_powergrid}} is provided, including a series of Jupyter notebooks designed to assist participants in getting started with the power flow simulation and explaining the submission procedure. It also featured a fully documented implementation of the baseline solution, aiding participants in replicating baseline results.

There was a dedicated website for the competition  \footnote{\href{https://ml-for-physical-simulation-challenge.irt-systemx.fr/powergrid-challenge/}{https://ml-for-physical-simulation-challenge.irt-systemx.fr/powergrid-challenge/}}. It aimed to serve as a central hub for all essential information, including:
\begin{enumerate*}[label=(\roman*)]
\item General details;
\item Organization, rules, and regulations;
\item Links and instructions for the submission platform (Codabench);
\item Announcements and recordings of webinars;
\item Tutorial section with links to relevant parts of the starting kit repository;
\item Contact information for the organizers.
\end{enumerate*}

Interactive webinars were also conducted and recorded to offer insights into the competition, power flow simulation, and submission process on Codabench. Additionally, we have set up a dedicated email address and Discord channel to facilitate open communication with all participants (or potential participants). During the preliminary edition of the challenge, assistance and information were provided via Discord, and the organizers remained available to help all participants.

\section{Competition results and winning solutions}
\label{sec:outcomes_participants}
This section presents and explores the results of the competition. We present the overall leaderboard with all the participants, a detailed analysis of winning solutions from the perspective of four categories of evaluation criteria, and provide the description of their methodology.

\subsection{Submission performance overview}
An ensemble of 80 participants have joined the competition on Codabench during the different phases of the competition. The participants were representing whether the teams or individuals with diverse nationalities and were mainly from academia. At the end of the competition, there were 8 distinct entities in the final leaderboard on Codabench\footnote{\url{https://www.codabench.org/competitions/2378/\#/results-tab}}, as shown in Table \ref{tab: benchmark_codabench}. The ranking is based on the global score which represents an aggregation of other sub-scores (ML, physics, OOD, and speed-up). Two submissions were eliminated from the ranking, as they have only optimized their solutions for speed-up while neglecting completely other evaluation criteria categories.  

\begin{table}[H]
\caption{Competition Benchmark table with all the participants. The benchmark table is similar to the one available on Codabench, where the results are obtained over on execution of each model. The solution ranked third in Codabench presents a score $0.0$ from ML point of view and is eliminated from the competition. A more detailed analysis of first three valid solutions is provided in Table \ref{tab:Benchmark_evaluation_circle}.}
\resizebox{\linewidth}{!}{
\begin{tabular}{ccc|cccccc}
\hline
\cellcolor[HTML]{C0C0C0}           & \cellcolor[HTML]{C0C0C0} & \cellcolor[HTML]{C0C0C0} & \multicolumn{6}{c}{\textbf{Criteria categories}}                                                                                                    \\ \hline
\multicolumn{1}{c|}{\textbf{Rank}} & \multicolumn{1}{c|}{\textbf{Participant}} & \textbf{From}     & \multicolumn{1}{c|}{\textbf{Global score}} & \textbf{ML - test dataset} & \textbf{Physics - test dataset} & \textbf{ML - OOD} & \textbf{Physics OOD} & \textbf{Speed-up} \\ \hline
\multicolumn{1}{c|}{1}             & \multicolumn{1}{c|}{imshabansatti}          & USA             & \multicolumn{1}{c|}{62.88}               & 0.66                       & 0.28                            & 0.66              & 0.28                 & 0.17              \\
\multicolumn{1}{c|}{2}             & \multicolumn{1}{c|}{XJTU}         & China             & \multicolumn{1}{c|}{57.96}                & 0.66                       & 0.17                            & 0.55              & 0.17                 & 0.29              \\
\multicolumn{1}{c|}{-}             & \multicolumn{1}{c|}{\sout{MPData}}    & \sout{France}               & \multicolumn{1}{c|}{\sout{47.65}}                 & \sout{0.0}                        & \sout{0.13}                            & \sout{0.0}               & \sout{0.13}                 & \sout{1.0}               \\
\multicolumn{1}{c|}{3}             & \multicolumn{1}{c|}{Cordyceps}         & Canada         & \multicolumn{1}{c|}{41.79}                 & 0.44                       & 0.11                            & 0.22              & 0.11                 & 0.39              \\
\multicolumn{1}{c|}{4}             & \multicolumn{1}{c|}{Patrick}         &  France          & \multicolumn{1}{c|}{35.83}                 & 0.11                       & 0.13                            & 0.11              & 0.13                 & 0.54              \\
\multicolumn{1}{c|}{5}             & \multicolumn{1}{c|}{SrRiverar}     & Colombia           & \multicolumn{1}{c|}{35.22}                 & 0.33                       & 0.11                            & 0.22              & 0.09                 & 0.32              \\
\multicolumn{1}{c|}{6}             & \multicolumn{1}{c|}{Andyuoe}         &    N/A       & \multicolumn{1}{c|}{34.47}                 & 0.11                       & 0.13                            & 0.11              & 0.13                 & 0.51              \\
\multicolumn{1}{c|}{7}             & \multicolumn{1}{c|}{MPData}       & France             & \multicolumn{1}{c|}{30.79}                 & 0.33                       & 0.11                            & 0.11              & 0.09                 & 0.3               \\
\multicolumn{1}{c|}{-}             & \multicolumn{1}{c|}{\sout{Kuldeep}}       & \sout{India}           & \multicolumn{1}{c|}{\sout{25.47}}                 & \sout{0.0}                        & \sout{0.11}                            & \sout{0.0}               & \sout{0.09}                 & \sout{0.49}              \\ \hline
\end{tabular}}
\label{tab: benchmark_codabench}
\end{table}

Three solutions among the different submissions (ignoring the one which has been eliminated) obtained higher scores than the baseline which was based on LEAPNet architecture \cite{donon2020leap} (with a global score of 37.69) within a training and inference budget fixed to 12 hours per submission. We have noticed a diversity of the approaches proposed by different participants. They have used mainly the neural networks based architectures like graph neural networks, LeapNets, transformers and fully connected networks. Some participants have enhanced the baseline models by addressing their limitations. For example, the solution ranked second enhanced the LeapNet architecture to better adhere to fundamental physical laws. Section \ref{sec:res_details} provides an in-depth analysis of the top three winning solutions, after their description in the following section.

\subsection{Summary of submissions and description of top 3 methods}
This section provides a detailed description of the top three surrogate models that performed best in the simulation of power flow.

\subsubsection{HyPowerFlow: Hypersparse Message Passing for Parallel Power Flow on GPUs (Arizona State University, 1st Place)}
\begin{figure}[ht]
    \centering
\includegraphics[width=1.0\textwidth]{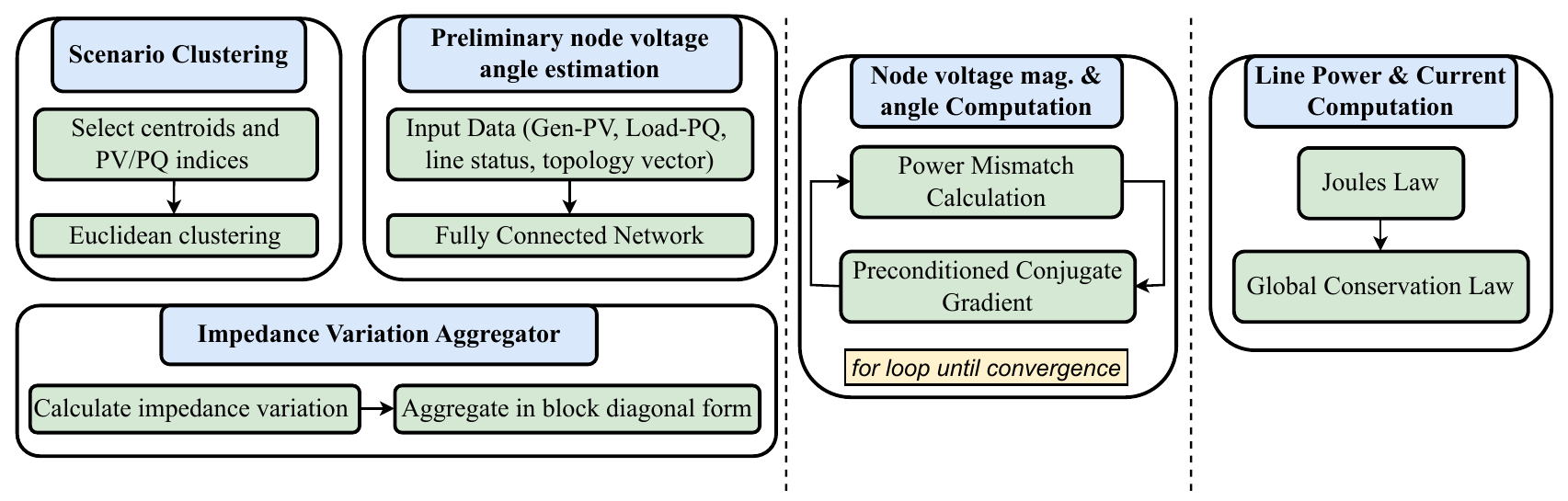}
    \caption{Overview of the HyPowerFlow Algorithm.}
\label{fig:HyPowerFlow}
\end{figure}
The HyPowerFlow\footnote{\url{https://github.com/amar-asu/HyPowerFlow}} algorithm in Figure~\ref{fig:HyPowerFlow} accelerates power flow convergence across multiple grid scenarios by leveraging GPU parallelism, data compression, structured hyper-sparsity, and refined initial estimates. Efficient data formatting optimizes GPU utilization and eliminates redundant matrix re-creation \cite{zeng2021gpu}. Structured hyper-sparsity exploits redundancy in the storage-intensive matrices to minimize memory overhead \cite{zlatev2013computational}. Improved initial estimate of node voltage angles reduces iterations and accelerates convergence.

Admittance matrix in power grid varies with topology changes across scenarios. Storing a separate matrix for each scenario is memory-intensive despite its sparsity. HyPowerFlow addresses this by analyzing impedance variations in a scenario, storing only localized modifications to a base admittance matrix instead of storing a scenario-specific admittance matrix. The base admittance matrix is a dense matrix common to all scenarios. A hyper-sparse delta admittance matrix captures sparse differences per scenario. The delta admittance matrices for all scenarios are aggregated into a block diagonal matrix \cite{zhang2013localized}.

HyPowerFlow employs clustering to avoid redundant matrix recreation. Different grid topologies are clustered together based on structural similarities \cite{zhou2009graph}. A consistent generator and load bus index is maintained across all scenarios in a cluster. This consistency simplifies the application of block diagonal matrices, allowing them to be used effectively in solving multiple contingency cases together. This structured approach simplifies computation and enhances scalability.

A fully connected neural network (FCN) is incorporated in HyPowerFlow to accelerate the power flow calculation process. Rather than relying on traditional power flow initialization methods, the FCN predicts an initial estimate of voltage angles based on power system variables. This improves the convergence of the power flow solution , complementing the efficiency gained through sparse matrix techniques and GPU acceleration.

An optimized Preconditioned Conjugate Gradient (PCG) method \cite{adabag2024mpcgpu} is, then, employed to solve the large and sparse linear systems involved in power flow solution \cite{stott2007fast}. By applying sparse matrix techniques and diagonal preconditioning, system equations are solved efficiently without explicitly storing large matrices. The delta admittance matrix approach is also extended to the Jacobian matrix in this step, further reducing computation overhead. The hyper-sparse representation of the Jacobian and admittance matrices for multiple scenarios enables us to solve the power flow for all these scenarios in parallel.

Finally, power and current calculations are performed using Joule’s law and global convergence principles. These calculations are also processed in parallel on the GPU, taking advantage of its ability to handle large-scale matrix operations efficiently \cite{zhang2022parallel}. The integration of physics-based constraints ensures the reliability and explainability of power flow solution.

A key feature of HyPowerFlow is its reliance on sparse matrix-vector multiplication, which significantly optimizes power flow analysis. Since delta matrices contain only few nonzero elements, storing them in a compressed format reduces memory consumption and avoids unnecessary calculations. The block diagonal structure further supports large-scale simulations with minimal computational overhead.

The structured hyper-sparsity of matrices in HyPowerFlow makes the approach well-suited for GPU acceleration. The compressed storage format ensures that only relevant data is retained, significantly reducing the memory footprint required for large-scale power system simulations. This is especially important for real-time applications, where handling a large number of sparse matrices would be impractical due to excessive memory demands.

\subsubsection{LEAP-PINN: Machine Learning for Solving Physical Simulations based on LEAPNet and KKT-hPINN (Xi’an Jiaotong University, 2nd place)}
\label{subsubsec:leap-pinn}
\begin{figure}[ht]
    \centering
\includegraphics[width=0.95\textwidth]{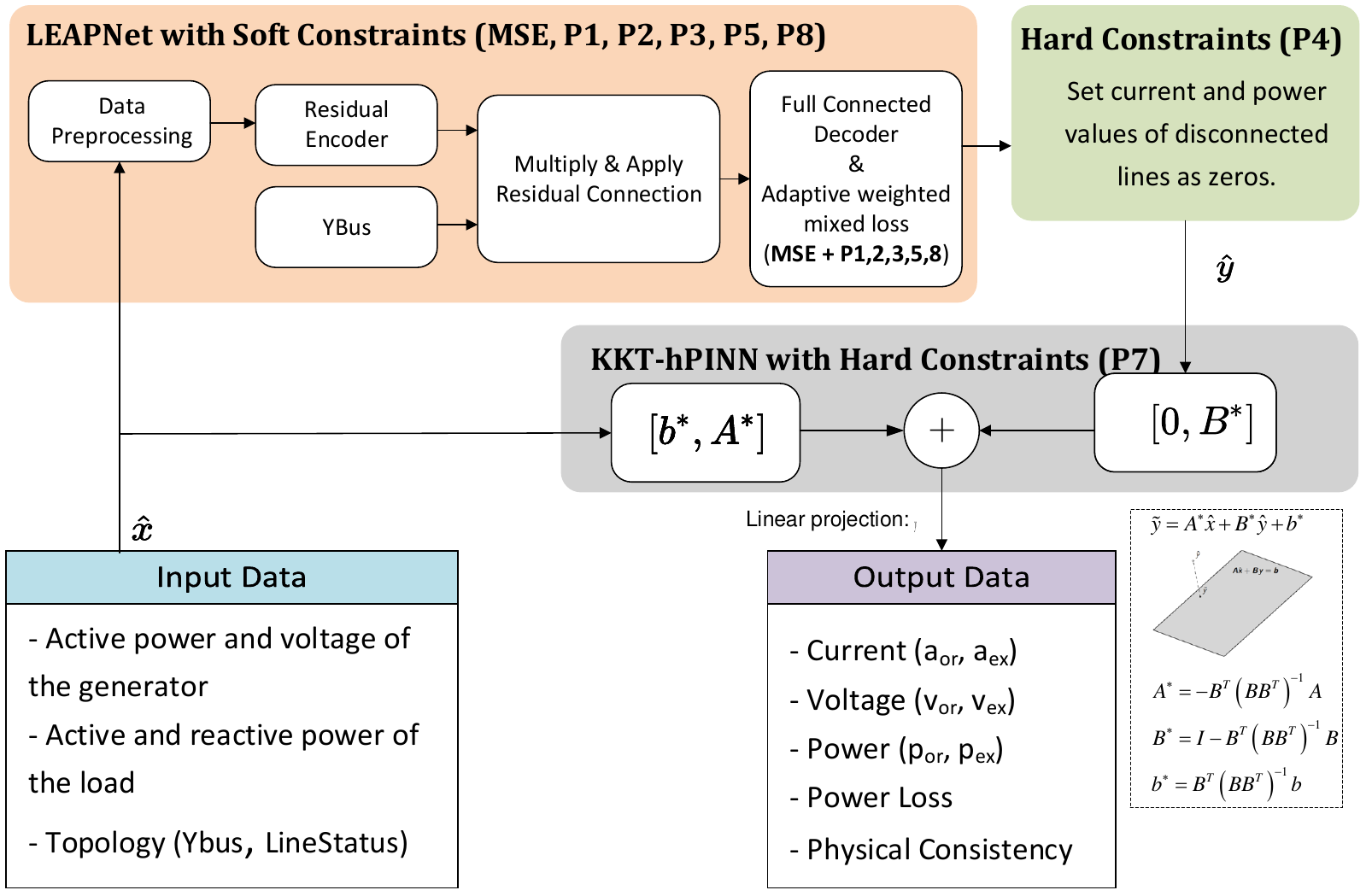}
    \caption{Overview of the LEAP-PINN architecture.}
\label{fig:leap_pinn_workflow_detailed}
\end{figure}

As shown in Figure~\ref{fig:leap_pinn_workflow_detailed}, the LEAP-PINN\footnote{\url{https://github.com/JiangJiang65/LEAP-PINN}} method presents a novel physics-informed neural network (PINN) architecture for efficient handling of power flow simulations.
LEAP-PINN integrates the capabilities of LEAPNet, a network architecture supporting topology changing \cite{donon2020leap}, with hard constraints derived from Karush–Kuhn–Tucker (KKT) conditions (referred to as KKT-hPINN) \cite{chen2024physics}. The detailed workflow of LEAP-PINN is outlined below:
\begin{itemize}
\item \textit{Soft constraints (P1, P2, P3, P5, P8) by LEAPNet:} Firstly, the input data are sent to the LEAPNet module, where training occurs with soft constraints integrated alongside topological insights. To enhance OOD generalization, LEAPNet explicitly captures variations in network topology by incorporating tensorized admittance matrices YBus. A hybrid loss function is employed, merging Mean Squared Error (MSE) with physics-driven penalty terms to ensure that the model not only learns from the data but also respects fundamental physical laws. The total loss function is defined as $\text{TotalLoss} = 100 \cdot \text{MSE} + \sum_{i=1,2,3,5,8} w_i \cdot P_i$, where constraints P1 and P2 enforce non-negativity for line currents ($a_{or}, a_{ex}$) and voltages ($v_{or}, v_{ex}$) by employing binary penalties. 
Constraints P3 and P5 penalize negative power losses ($p_{or} + p_{ex}$) and ensure that energy losses fall within the range of $[0.005, 0.04]$ of total production. Additionally, P8 indicates Joule's law compliance. 
With this loss function, LEAP-PINN allows the model to adapt to various grid topology configurations while maintaining physical consistency as far as possible.
   
\item \textit{Hard constraints (P4):} Secondly, we address constraint P4 by explicitly assigning power and current values of disconnected lines to zero. This ensures the rigorous fulfillment of constraint P4.

\item \textit{Hard constraints (P6, P7) by KKT-hPINN:} Thirdly, we address the enforcement of global conservation (P6) and local conservation (P7) of energy. Since P6 can be deduced from P7, our focus here lies on P7. To ensure strict adherence of the predicted results $\hat{y}$ from the previous step to P7, we project $\hat{y}$ onto the linear hyperplane spanned by P7. Specifically, leveraging the KKT-hPINN method \cite{chen2024physics}, we obtain $\tilde{y}$ from $\hat{y}$ in accordance with the KKT conditions \cite{boyd2004convex}.
\end{itemize}

\subsubsection{Transformer-based solution to solve power flow (University of Toronto, 3rd place)}

\begin{figure}[ht]
    \centering
    \includegraphics[width=0.95\textwidth]{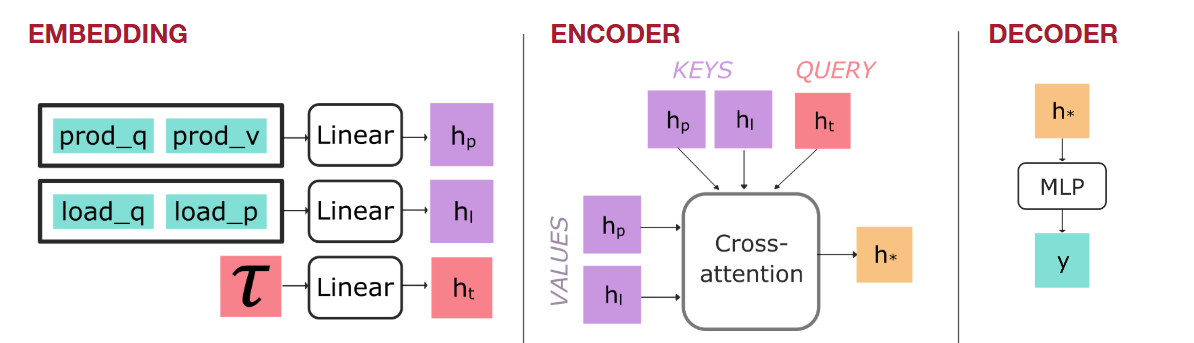}
    \caption{Overview of the cross-attention based architecture.}
    \label{fig:transformer-overview}
\end{figure}

Figure \ref{fig:transformer-overview} gives an overview of our framework\footnote{\url{https://github.com/Ninalgad/powergrid-self-attention}}. In summary, we compute embeddings of the input variables where the \textit{topology} embedding attends to the grid state representations (cross-attention) to predict the target output. 
Here, the cross-attention module is a multihead attention \cite{vaswani2017attention} mechanism defined as:
\begin{equation} \label{eq1}
\begin{split}
\text{MultiHead(Q,K,V)} &= \text{concat}(\text{head}_1,..,\text{head}_H)W \\
\text{where } \text{head}_i &= \text{Attention}(QW_i^Q, KW_i^K, VW_i^V) \\
\text{where } \text{Attention}(Q^\prime,K^\prime,V^\prime) &= \text{softmax}(Q^\prime K^{\prime T}/\sqrt{d_k})V^\prime
\end{split}
\end{equation}

With the weight matrices $W\in\mathbb{R}^{hd_v\times d_{model}}$, $W_i^Q\in\mathbb{R}^{d_{model}\times d_k}$, $W_i^K\in\mathbb{R}^{d_{model}\times d_k}$ and $W_i^V\in\mathbb{R}^{d_{model}\times d_v}$.
Our model uses $H=8$ heads with embedding dimensions $d_v=d_k=d_{model}=512$. Note that we did not add additional attention layers to minimize the training and inference times. 

In detail, we reduce the number of vectors given to the attention module by grouping vectors that encode similar information to reduce the computational burden introduced by the attention operation.
We then create a single embedding vector for each group, creating a sequence of embedding vectors. 
The resulting sequence is then fed to the cross-attention layer to model the interactions between these groupings and determine which input variables are relevant for a given topology query. 
This resulting query-specific representation $h_*$ is then decoded by an MLP to predict the target variables $y$.

\subsection{Results details and discussion}
\label{sec:res_details}

Table \ref{tab:Benchmark_evaluation_circle} compares the three winning solutions with provided baselines. It uses a synthetic visualization of the performances using their discretization based on thresholds, a procedure that is described in \ref{app:thresholds}. This facilitates the reading of the table and is also used for the computation of the global score. To ensure the robustness of the benchmark, we have executed each solution 10 times (training and evaluation) and the results are represented using average performances and standard deviation is also reported for the global score. The first row of this table presents the physical simulator based on lightsim2grid solver. It presents great performance (green circle) for all the metrics as the physical solvers provide the exact solutions to the problem and is also used as the ground truth. Concerning the speed-up metric, we compare the performance of all the solutions with this baseline and hence it is set to $1$. We have also provided two surrogate baselines for this competition, which were based on Fully Connected (MLP) and LeapNet architectures. These baselines accelerate the physical solver computation at the cost of large losses over other evaluation criteria (accuracy and physics compliance). They have obtained a global score of 33\% and 37\% respectively, and which is also set the minimum score required to be eligible for the competition prizes. Finally, the solutions submitted by the winners of the competition are ranked and their performances are analyzed in the following. 

\begin{table}[htbp]
    \centering
    \caption{Competition Benchmark Table - Scoring table for load flow prediction task under 3 categories of evaluation criteria. The performances are reported using three colors computed on the basis of two thresholds. Colors meaning: \protect\myfullcircle{red} Unacceptable (0 point) \protect\myfullcircle{orange} Acceptable (1 point)  \protect\myfullcircle{green}  Great (2 points). Reported results: Ground Truth (Powerflow), Accelerated physical solver (Security Analysis), Two baselines based on Neural Networks (Fully Connected and LeapNet architectures). The ranking is : 1) Arizona State University (ASU), 2) Xi'an Jiaotong University (XJTU), 3) University of Toronto (UToronto).}
    \resizebox{\columnwidth}{!}{
    \begin{tabular}{cccccccccc}
    \toprule
         & \multicolumn{8}{c}{\textbf{Criteria category}}\\
         & \multicolumn{2}{c}{\textbf{Test dataset (30\%)}} && \multicolumn{1}{c}{\textbf{Speed-up (40\%)}} && \multicolumn{2}{c}{\textbf{OOD generalization (30\%)}}  && \textbf{Global}\\ \cline{2-3} \cline{5-5} \cline{7-8} 
         \textbf{Method} & ML-related(66\%) & Physics(34\%) && Speed-up &&  ML-related(66\%) & Physics(34\%) && \textbf{Score (\%)}\\ \cline{1-10}
         
         \multicolumn{1}{c|}{Powerflow} & \underline{$a_{or}$}\, \underline{$a_{ex}$}\, \underline{$p_{or}$}\,\, \underline{$p_{ex}$}\,\,\, \underline{$v_{or}$}\,\, \underline{$v_{ex}$} & \underline{$p_1$}\, \underline{$p_2$}\, \underline{$p_3$}\,\, \underline{$p_4$}\,\, \underline{$p_5$}\,\, \underline{$p_6$}\,\, \underline{$p_7$}\,\, \underline{$p_8$} && &&  \underline{$a_{or}$}\, \underline{$a_{ex}$}\, \underline{$p_{or}$}\,\, \underline{$p_{ex}$}\,\,\, \underline{$v_{or}$}\,\, \underline{$v_{ex}$} & \underline{$p_1$}\, \underline{$p_2$}\, \underline{$p_3$}\,\,\, \underline{$p_4$}\,\, \underline{$p_5$}\,\, \underline{$p_6$}\,\, \underline{$p_7$}\,\, \underline{$p_8$} &&\\

        \multicolumn{1}{c|}{(LightSim2grid)} & \hspace{-0.1cm}\myfullcircle{green}\,\,\myfullcircle{green}\,\,\,\,\myfullcircle{green}\,\,\,\myfullcircle{green}\,\,\,\,\myfullcircle{green}\,\,\,\myfullcircle{green} & \myfullcircle{green}\myfullcircle{green}\myfullcircle{green}\myfullcircle{green}\myfullcircle{green}\myfullcircle{green}\,\myfullcircle{green}\,\myfullcircle{green} && 1 &&  \hspace{-0.1cm}\myfullcircle{green}\,\,\myfullcircle{green}\,\,\,\,\myfullcircle{green}\,\,\,\myfullcircle{green}\,\,\,\,\myfullcircle{green}\,\,\,\myfullcircle{green} & \myfullcircle{green}\myfullcircle{green}\myfullcircle{green}\myfullcircle{green}\myfullcircle{green}\myfullcircle{green}\,\myfullcircle{green}\,\myfullcircle{green} && \textbf{60.2}\\

        \multicolumn{1}{c|}{(Security analysis)} & \hspace{-0.1cm}\myfullcircle{green}\,\,\myfullcircle{green}\,\,\,\,\myfullcircle{green}\,\,\,\myfullcircle{green}\,\,\,\,\myfullcircle{green}\,\,\,\myfullcircle{green} & \myfullcircle{green}\myfullcircle{green}\myfullcircle{green}\myfullcircle{green}\myfullcircle{green}\myfullcircle{green}\,\myfullcircle{green}\,\myfullcircle{green} && 3.77 &&  \hspace{-0.1cm}\myfullcircle{green}\,\,\myfullcircle{green}\,\,\,\,\myfullcircle{green}\,\,\,\myfullcircle{green}\,\,\,\,\myfullcircle{green}\,\,\,\myfullcircle{green} & \myfullcircle{green}\myfullcircle{green}\myfullcircle{green}\myfullcircle{green}\myfullcircle{green}\myfullcircle{green}\,\myfullcircle{green}\,\myfullcircle{green} && \textbf{62.5}\\
        \hline

         \multicolumn{10}{c}{\textbf{Competition baselines}} \\\hline
         
         \multicolumn{1}{c|}{Fully Connected} & \hspace{-0.1cm}\myfullcircle{orange}\,\,\myfullcircle{orange}\,\,\,\,\myfullcircle{red}\,\,\,\myfullcircle{red}\,\,\,\,\myfullcircle{orange}\,\,\,\myfullcircle{orange} & \myfullcircle{green}\myfullcircle{green}\myfullcircle{red}\myfullcircle{red}\myfullcircle{orange}\myfullcircle{red}\,\myfullcircle{red}\,\myfullcircle{red} && 15.45 &&  \hspace{-0.1cm}\myfullcircle{red}\,\,\myfullcircle{red}\,\,\,\,\myfullcircle{red}\,\,\,\myfullcircle{red}\,\,\,\,\myfullcircle{red}\,\,\,\myfullcircle{red} & \myfullcircle{green}\myfullcircle{green}\myfullcircle{red}\myfullcircle{red}\myfullcircle{orange}\myfullcircle{red}\,\myfullcircle{red}\,\myfullcircle{red} && \textbf{33.5}\\

         \multicolumn{1}{c|}{LeapNet} & \hspace{-0.1cm}\myfullcircle{green}\,\,\myfullcircle{green}\,\,\,\,\myfullcircle{orange}\,\,\,\myfullcircle{orange}\,\,\,\,\myfullcircle{red}\,\,\,\myfullcircle{red} & \myfullcircle{green}\myfullcircle{green}\myfullcircle{red}\myfullcircle{red}\myfullcircle{orange}\myfullcircle{red}\,\myfullcircle{red}\,\myfullcircle{red} && 11.9 &&  \hspace{-0.1cm}\myfullcircle{orange}\,\,\myfullcircle{orange}\,\,\,\,\myfullcircle{orange}\,\,\,\myfullcircle{orange}\,\,\,\,\myfullcircle{red}\,\,\,\myfullcircle{red} & \myfullcircle{green}\myfullcircle{green}\myfullcircle{red}\myfullcircle{red}\myfullcircle{orange}\myfullcircle{red}\,\myfullcircle{red}\,\myfullcircle{red} && \textbf{37.6}\\
         \hline \hline
        
         \multicolumn{10}{c}{\textbf{Competition Ranking}} \\\hline
         
         \multicolumn{1}{l|}{1-ASU} & \hspace{-0.1cm}\myfullcircle{green}\,\,\myfullcircle{green}\,\,\,\,\myfullcircle{green}\,\,\,\myfullcircle{green}\,\,\,\,\myfullcircle{green}\,\,\,\myfullcircle{green} & \myfullcircle{green}\myfullcircle{green}\myfullcircle{green}\myfullcircle{green}\myfullcircle{orange}\myfullcircle{green}\,\myfullcircle{red}\,\myfullcircle{green} && 7.87 &&  \hspace{-0.1cm}\myfullcircle{green}\,\,\myfullcircle{green}\,\,\,\,\myfullcircle{green}\,\,\,\myfullcircle{green}\,\,\,\,\myfullcircle{green}\,\,\,\myfullcircle{green} & \myfullcircle{green}\myfullcircle{green}\myfullcircle{green}\myfullcircle{green}\myfullcircle{orange}\myfullcircle{green}\,\myfullcircle{red}\,\myfullcircle{green} && \textbf{$64.2 \pm .62$}\\
         \hline

         \multicolumn{1}{l|}{2-XJTU} & \hspace{-0.1cm}\myfullcircle{green}\,\,\myfullcircle{green}\,\,\,\,\myfullcircle{orange}\,\,\,\myfullcircle{orange}\,\,\,\,\myfullcircle{green}\,\,\,\myfullcircle{green} & \myfullcircle{green}\myfullcircle{green}\myfullcircle{red}\myfullcircle{red}\myfullcircle{orange}\myfullcircle{orange}\,\myfullcircle{green}\,\myfullcircle{red} && 9.69 &&  \hspace{-0.1cm}\myfullcircle{green}\,\,\myfullcircle{green}\,\,\,\,\myfullcircle{orange}\,\,\,\myfullcircle{orange}\,\,\,\,\myfullcircle{green}\,\,\,\myfullcircle{green} & \myfullcircle{green}\myfullcircle{green}\myfullcircle{red}\myfullcircle{red}\myfullcircle{orange}\myfullcircle{orange}\,\myfullcircle{green}\,\myfullcircle{red} && \textbf{$57.89 \pm 1.42$}\\
         \hline
         
         \multicolumn{1}{l|}{3-UToronto} & \hspace{-0.1cm}\myfullcircle{green}\,\,\myfullcircle{green}\,\,\,\,\myfullcircle{green}\,\,\,\myfullcircle{green}\,\,\,\,\myfullcircle{red}\,\,\,\myfullcircle{red} & \myfullcircle{green}\myfullcircle{green}\myfullcircle{red}\myfullcircle{red}\myfullcircle{orange}\myfullcircle{red}\,\myfullcircle{red}\,\myfullcircle{red} && 12.42 &&  \hspace{-0.1cm}\myfullcircle{orange}\,\,\myfullcircle{orange}\,\,\,\,\myfullcircle{orange}\,\,\,\myfullcircle{orange}\,\,\,\,\myfullcircle{red}\,\,\,\myfullcircle{red} & \myfullcircle{green}\myfullcircle{green}\myfullcircle{red}\myfullcircle{red}\myfullcircle{orange}\myfullcircle{red}\,\myfullcircle{red}\,\myfullcircle{red} && \textbf{$41.15 \pm 1.27$}\\
        \bottomrule
    \end{tabular}}
    \label{tab:Benchmark_evaluation_circle}
\end{table}

The first solution obtains a very promising result through the different categories of evaluation criteria. They achieved a slightly higher score with respect to the physical solver baseline, highlighting the potential of hybrid approaches to replace physical solvers in the future. It also shows similar behavior for both test and OOD datasets, emphasizing their generalization capacity. Due to the complexity of their approach in which the physical constraints are solved directly based on an optimization algorithm, the speed-up gain in not as high as expected, but it remains still a good candidate for further improvements. The second solution, suggesting an improvement over the LeapNet architecture by considering the physical constraints, obtained also a very similar performance and global score. They were able to improve considerably precision of LeapNet baseline in terms of ML-related accuracy (for voltages) by adding the physical constraints into the loss function. Their specific technique based on the projection of predictions onto the linear hyperplane spanned by local conservation law, improves its compliance for $P_6$ and $P_7$. Finally, the third solution which used a data-driven approach, without considering the physical constraints, obtained a slightly higher performance than the provided baselines and confirming the fact there may still be room for further improvement using physics-free solutions.

\begin{figure*}[t!]
    \centering
    \begin{subfigure}[t]{0.49\textwidth}
        \centering
        \includegraphics[scale=0.5]{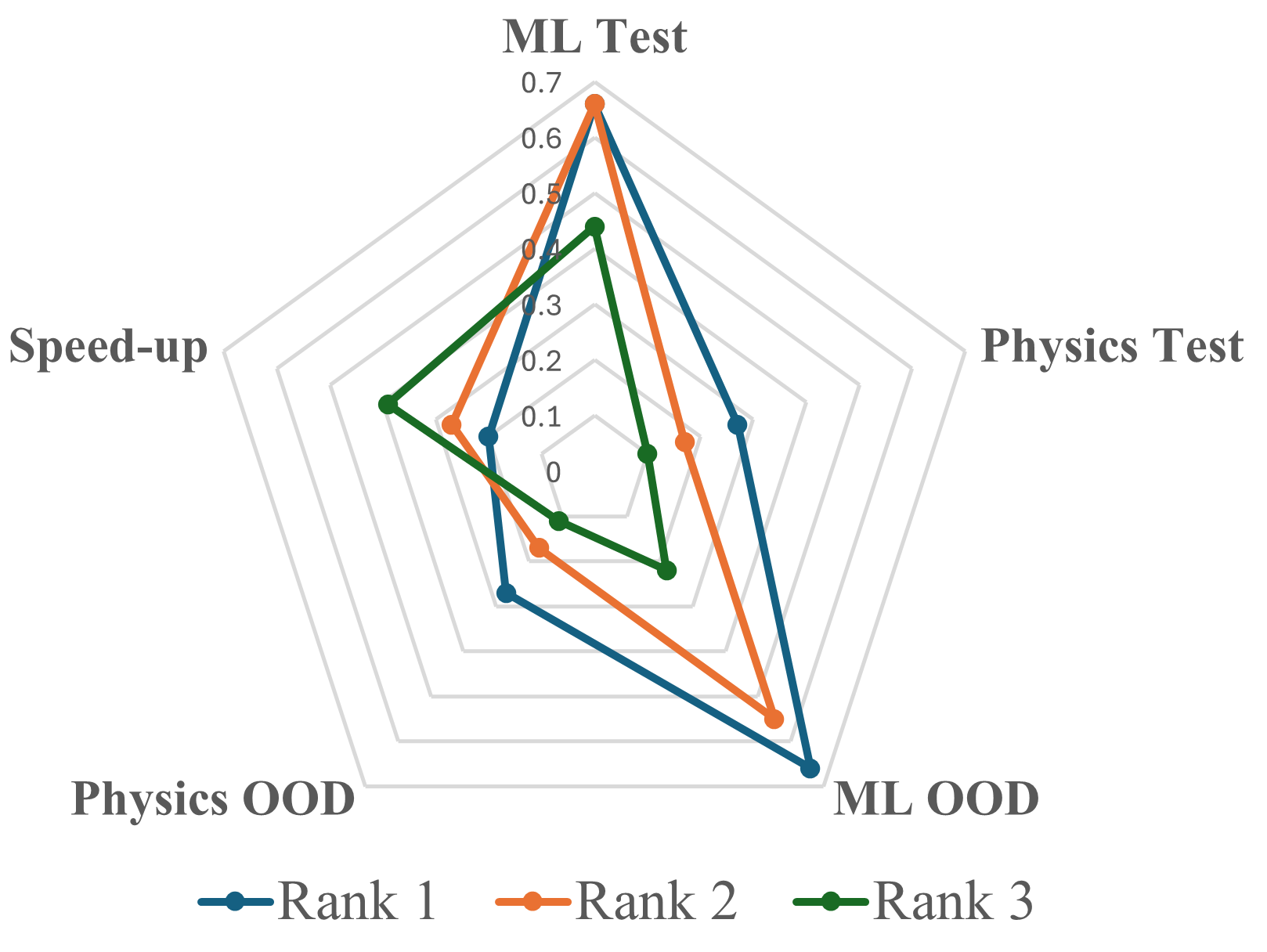}
        \caption{Sub-scores}
        \label{subfig:radar_chart_performance}
    \end{subfigure}%
    ~ 
    \begin{subfigure}[t]{0.49\textwidth}
        \centering
        \includegraphics[scale=0.5]{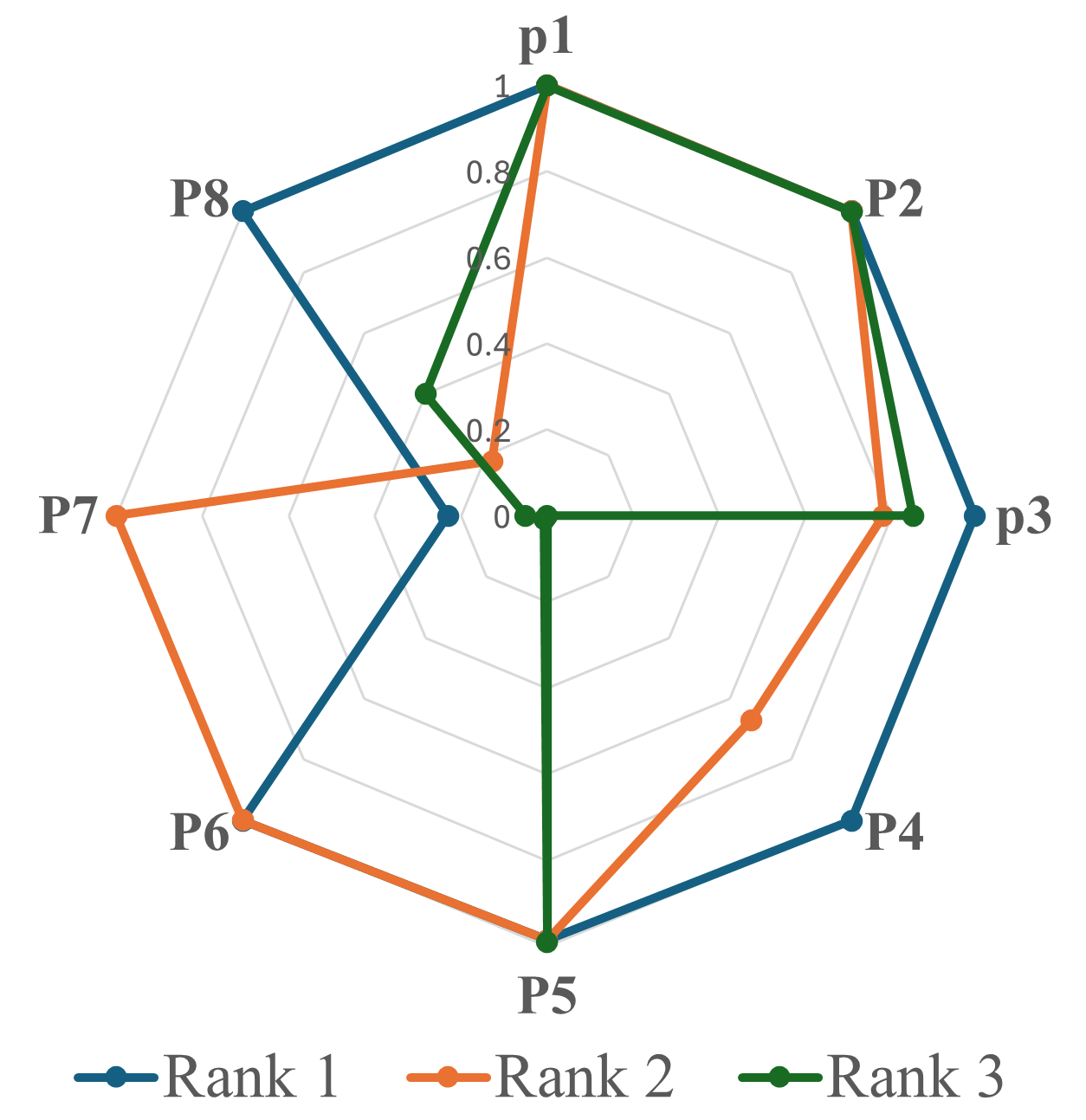}
        \caption{Physics criteria on test dataset}
        \label{subfig:radar_test}
    \end{subfigure}
    \caption{Radar chart presenting the performance of winning solutions using the eight physics criteria. The reported values ranging $[0,1]$ are the proportions of the compliant samples ($1 - \textnormal{violation}$). The physics criteria are: (P1) current positivity, (P2) voltage positivity, (P3) losses positivity, (P4) disconnected lines, (P5) energy loss, (P6) global conservation, (P7) local conservation, (P8) Joule law.}
    \label{fig:radar_physics}
\end{figure*}

To facilitate the comparison of winning solutions at a macro-level, their scores corresponding to each category of evaluation criteria are visualized using a radar chart in Figure \ref{subfig:radar_chart_performance}. We can notice a similar pattern among the first and second solutions with a slightly higher score obtained by the first solution for all the categories, except for the speed-up. It highlights also the potential to obtain higher speed-up using data-driven solutions (rank 3). It can also be noticed that, even if these approaches integrates the physical constraints in their design, their performances in terms of physics compliance remains still an avenue for further explorations and improvements.

To reduce the bias and smoothing effect due to the use of discretized metrics, in Figure \ref{subfig:radar_test}, we provide the proportion of the samples satisfying the eight physics laws obtained on \textit{test dataset}. The compliance to physics criteria remains very similar throughout both test and OOD datasets, confirming the physics compliance invariance of designed models to data distribution variations. This visualization may help to notice if two scores are very close, despite of being discretized into two different modalities. As an example, considering $P_3$ law, the performances for second and third solutions remain close to the first solution; however, they have been attributed a not-acceptable score (red color: 0 points) for this particular metric. Such observation may guide the further adjustment of some thresholds for future editions of this challenge. The first and second solutions show a very competitive performance on physics compliance. Their major differences reside on the violation of physics criteria $P_7$ and $P_8$, which are local conservation and Joule law respectively. This can be explained by the fact that the second solution has made use of a specific technique to make their approach compliant to local conservation law, whereas, the focus of the first solution was based on the Joule law at the final step. The Rank $2$ solution shows also a lower performance with respect to $P_3$ and $P_4$ which are losses positivity and predictions for disconnected lines. This latter was also the case for the solution with Rank $3$. These results highlight the different improvement perspectives that could be reached by these approaches.

\section{Organizational outcomes}
\label{sec:outcomes_organization}
The competition ran on Codabench competition platform through the different competition phases. In the following, we review the different contributions of this competition, the lessons learned, and also provide recommendations for possible improvements.

\subsection{Effectiveness of Competition Materials}
To the best of our knowledge, this competition was the first open challenge dedicated to applying Machine Learning for Physical Simulation (ML4PhySim) in the power grid domain. In organizing the competition, we leveraged insights from previous challenges, including one focused on airfoil design \cite{yagoubi2024neurips} and another related to power grid control \cite{marot2021learning}.
The competition framed power grid management as a regression problem, allowing participants without prior expertise in the field to contribute using ML-based solutions. To support participants in their model development, we provided a comprehensive starting kit, including a set of example implementations. Additionally, our baseline models were purely ML-based, without incorporating physical constraints. Notably, the third-place winner utilized a transformer-based approach, highlighting the growing influence of this architecture across various AI domains.

To encourage the adoption of more advanced hybrid models, we introduced supplementary evaluation criteria that considered physics compliance. As a result, solutions based purely on ML were penalized under this category, reflecting the fact that conventional ML models are typically designed for accuracy without accounting for underlying physical principles.

During the warm-up phase, we also provided an example of a hybrid model based on graph neural networks, using a simplified configuration where data was generated via Direct Current (DC) approximation. This implementation not only inspired participants to explore physics-informed ML solutions but also served as an educational resource for those new to the field.

\subsection{Incentive Measures}
To facilitate solution submissions on Codabench, we provided multiple options for participants. Those with access to computational resources—or to avoid potential infrastructure issues—could submit pre-trained models or final metrics for scoring and ranking on the competition's intermediate leaderboard. For participants without access to computational resources, we allocated two clusters of eight GPUs throughout the competition. However, to be eligible for prizes, all final submissions had to be trained and evaluated on our servers.

To encourage the participation in this competition, a total of five prizes were considered. Three general prizes---3\;000, 2\;000, and 1\;000 euros---were considered for the top three winners. Additionally, two special prizes, each worth 1\;000 euros, were considered for most accurate model based only on machine learning performance metrics and for the best student solution. To enhance the competitiveness, the general and special prizes were non-cumulative---winning a general prize made participants ineligible for a special prize.

To ensure fair resource distribution, each team was allowed up to 10 submissions per day (with a maximum of 2 simultaneous submissions) and a total of 1,000 submissions per phase. There are multiple advantages of training the participants' solutions on our infrastructure, among which we can cite: avoiding an unfair advantage by using high-end GPUs or private clusters for the final solutions where all the models are trained under identical conditions, ensuring comparable results; ensuring that the results to be fully reproducible, as training and evaluation occur in a controlled environment and prevent hidden data leakage or external pre-training biases; encouraging the participants to optimize their training process in addition to the inference; avoiding unnecessary data transfers and large model file. 

Additionally, we organized a webinar to introduce the competition framework, explain the problem context, demonstrate the use of the starting kit and datasets, and guide participants through the Codabench submission process. This session was particularly beneficial for those new to power grid modeling, offering an end-to-end demonstration of competition mechanics.

\begin{figure}
    \centering
    \includegraphics[width=1\linewidth]{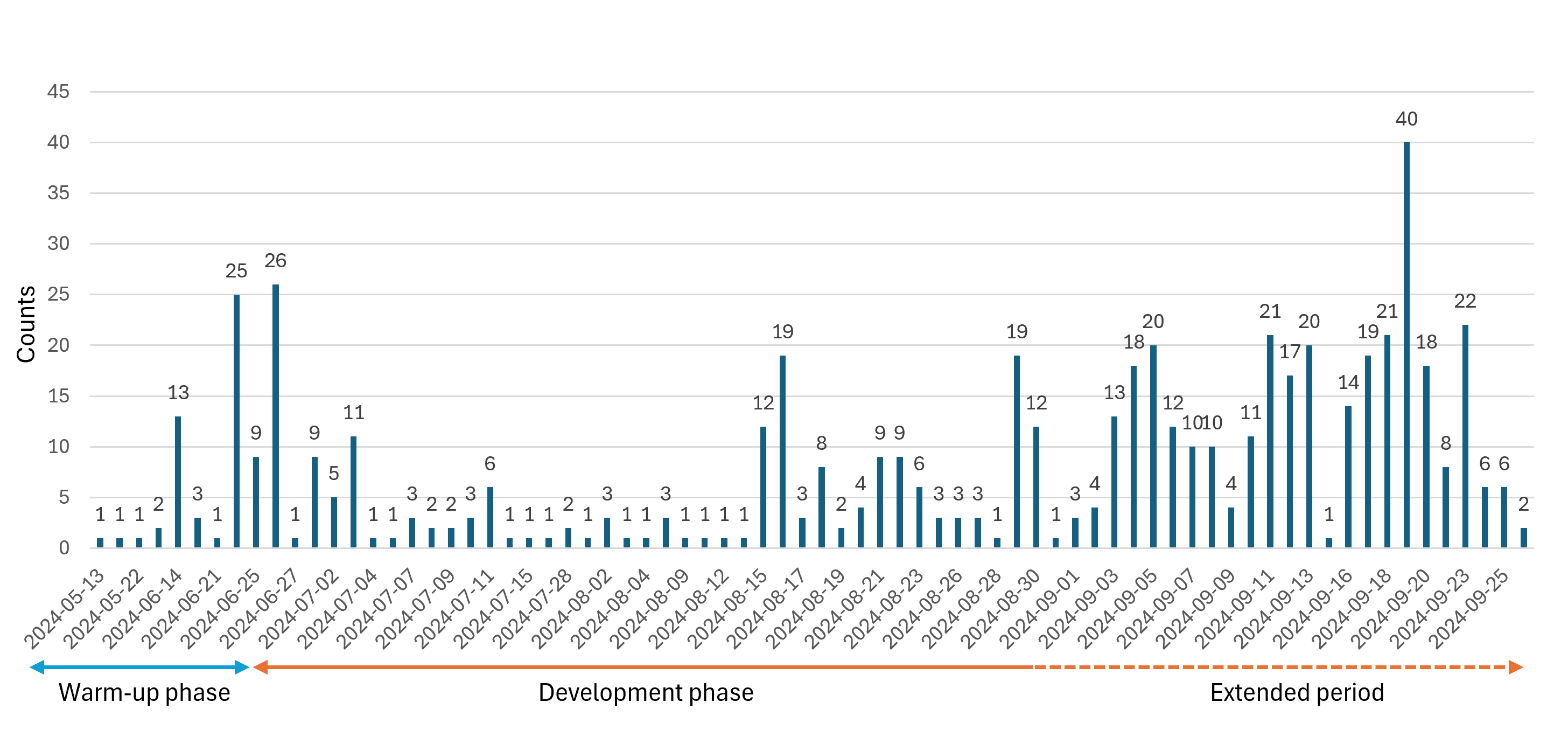}
    \caption{Distribution of submissions through competition phases and dates.}
    \label{fig:submission_dis}
\end{figure}

Figure \ref{fig:submission_dis} illustrates the submission distribution across different competition phases, showing three intervals of increased activity: one at the end of the warm-up phase, another near the initial conclusion of the development phase, and a final surge during the extended development period. The first surge coincided with the integration of additional libraries suggested by the participants, helping them better navigate the competition materials. The highest level of activity occurred during the three-week extension, underscoring the importance of providing extra time at the end of such challenges to accommodate last-minute submissions.

\subsection{Soundness of Evaluation Methodology}
Most challenges across various domains rely on a single evaluation criterion, such as model accuracy. In contrast, this competition employed a more comprehensive framework with four evaluation categories: accuracy, physics compliance, industrial readiness (speed-up), and out-of-distribution generalization. This multi-faceted approach encouraged participants to develop solutions that balanced multiple objectives, leading to a diverse set of methodologies. These included hybrid approaches that combined optimization algorithms with message passing and ML-based parameter initialization, augmented ML techniques that incorporated physical constraints as regularization terms, and purely ML-based solutions.

To assess generalization, we designed an additional dataset with a slightly different distribution from the standard test set, which is typically used in such evaluations. This approach allowed us to better differentiate solutions based on their real-world applicability, where the training and deployment distributions often do not perfectly align.

For scoring, we employed a discretization and weighting approach across the different evaluation categories. While this method was chosen for its simplicity and effectiveness in handling heterogeneous criteria, the results also highlighted certain nuances. Specifically, some solutions achieved similar scores within individual categories when measured using real-valued metrics. Although this had only a minor impact on the overall ranking, it is an aspect that could be refined in future iterations of the competition.

\subsection{Learning impact for participants and organizers: Simultaneous academic and industrial benefits}
The impact of the competition materials (e.g., starter kit, webinars) on participants was analyzed, not only in the challenge described here but also in a previous Machine Learning for Physical Simulation (ML4PHYSIM) challenge focused on an airfoil use case. Both competitions utilized the LIPS framework for evaluation and similar educational resources. In the following, we present key insights into the scientific, organizational, and educational impacts of the competition on both participants and organizers.

\paragraph{Scientific impact on participants}
The results show that, from a scientific point of view, the competition material can have various effects on the participants:
\begin{itemize}
    \item Some participants used the competition as an opportunity to validate and refine existing AI models. For them, the material did not significantly alter their scientific strategy but instead served to legitimize their approaches within their organizations;
    \item Others benefited from exposure to new scientific knowledge via the provided baseline models. This led to:
    \begin{itemize}
        \item Development of improved models, as seen in the case of the second-place winner who built upon a baseline;
        \item Enhanced understanding of hybrid or physical models and the powergrid use case for some participants.
    \end{itemize}
\end{itemize}

\paragraph{Organizational impact on participants}
From an organizational point of view, the following impacts are identified on the participants:
\begin{itemize}
    \item The competition served as a tool for internal recognition. Participants could highlight that their methods were independently evaluated and compared with a broad range of industrial and academic solutions, sometimes leading to additional internal support and resources;
    \item Especially for winners, the challenge also offered external visibility, boosting their reputation in the hybrid AI community.
\end{itemize}

\paragraph{Learning for organizers}
The challenge, grounded in energy transition scenarios and renewable integration, yielded valuable insights:
\begin{itemize}
    \item \textit{For TSO}: Some internal baseline models were outperformed, and new, untested model types proved effective. This validated the need to launch operational projects involving hybrid AI;
    \item \textit{For organizers}: The submissions brought fresh scientific perspectives. While the competition series (including the previous and upcoming airfoil challenges) offered rich insights, they also revealed a key limitation: solutions remain highly use-case specific. Generalization of hybrid AI models is a critical frontier.
\end{itemize}

In summary, the competition facilitated mutual learning. Participants gained scientific skills and visibility, while organizers obtained valuable feedback on real-world challenges in energy transition and hybrid AI research. This reflects a double industrial and academic impact, reinforcing the relevance of such competitions in fostering innovation \cite{gilain2025training, plantec2024simultaneous}.

\section{Conclusions and perspectives}
\label{sec:conclusions}
This paper presents our challenge called Machine Learning for Physical Simulation (ML4PhySim) dedicated to power flow simulation. The objective of this challenge was to undertake the first step towards accelerating, through the use of AI, the computationally intensive physical solvers used currently in TSO's operation rooms in different scenarios. It was organized following the successful challenges in this field that we have organized on another physical problem, and to best of our knowledge is the first competition that promotes the use of hybrid models for power grid simulation. The solutions submitted by different teams in this competition, shed light on the potential benefit of the use of ML-based approaches in power grid simulation, where the winning team's solution outperformed the physical solver performance. We have also noticed a range of heterogeneous approaches that have been suggested, such as physics informed neural networks, graph based approaches, constrained optimization and transformers. The unique evaluation framework based on LIPS enabled also the identification of weaknesses and strengths of the solutions and providing valuable insights for their further improvements.

\paragraph{Limitations and perspectives} In this challenge, the scenario was designed to reflect the real-world configuration in power systems, such as integration of renewable energies and allowing the topological changes in power grid. One insight for future competitions is to consider the scalability of the proposed approaches on the real data. Another direction on scalability could be relied on considering two grid sizes, where the models trained on the smaller grid should provide robust performances on a larger one. The considered industrial readiness evaluation category could also be extended to include the complexity and memory consumption of the models.

Based on the submitted solutions in this and similar challenges focused on modeling physical phenomena, we observed that many approaches exhibit context-specific characteristics. This suggests that significant adaptations may be required when the problem configuration changes. We believe that organizing more challenges with diverse scenarios and configurations will help drive the development of more generalized approaches, such as foundation models for power grids \cite{hamann2024perspective}---commonly referred to as Grid FM\footnote{\url{https://lfenergy.org/projects/gridfm/}}.

\paragraph{Societal impact} Another aspect was related to the promotion of the problems related to power sector. Through the organization of this challenge, we wanted to attract the attention of AI researchers from academia and industry to this specific problem, within the context of the globalization of the energy market and energy transition towards the renewable resources. We hope that the organization and the contuinty of such challenges would help to accelerate the research in the energy sector.

\section*{Acknowledgments}
This work has been supported by SystemX Technological Research Institute and in the context of HSA project (hybridization of simulation and ML). As part of HSA, RTE France (main TSO in France) provides its support through valuable insights concerning the competition use case and power grid systems. We are very thankful to NVIDIA France and Benoit Bastien for sponsoring through their infrastructure these developments by making GPU servers available for training all baselines as well as for running the Codabench public evaluation thread. Finally, we would like thank all the participants of the competition, and specifically the winners who have contributed, by providing a theoretical description of their approaches, in this paper.

\newpage
\appendix

\section{Notation table}
\label{app1}

The following table introduces the notations used throughout the paper and the Appendix.

\begin{table}[htb]
\caption{Notation table}
\scalebox{0.9}{
\begin{tabular}{l|l}
Symbol & Description \\ \hline
$i$ & \text{Index associated with an individual (entity)} \\
$t$ & \text{Index associated with a time instant} \\
$k$ & \text{Index associated with a grid node}\\
$\ell$ & \text{Index associated with a power line}\\
$or$ & \text{Index associated with the "origin" extremity of a power line by convention}\\
$ex$ & \text{Index associated with the second extremity of a power line by convention}\\
$j$ & \text{The imaginary part of complex numbers}\\
$p_k$ & \text{Active power at node $k$}\\
$q_k$ & \text{Reactive power at node $k$}\\
$v_k$ & \text{Voltage at node $k$}\\
$\theta_k$ & \text{Voltage phase (angle) at node $k$}\\
$p^\ell$ & \text{Active power flow over line $l$}\\
$q^\ell$ & \text{Reactive power flow over line $l$}\\
$a^\ell$ & \text{Current over line $l$}\\
$L$ & \text{Number of power lines in the network}\\
$K$ & \text{Number of substations in the network}\\
$N$ & \text{Number of samples in training set} \\
$M$ & \text{Number of samples in test set} \\
\end{tabular}}
\label{tbl: notation}
\end{table}

\newpage
\section{Power flow equations}
\label{app2}
For a given line $\ell$, the active and reactive ${p}^{\ell}$ and ${q}^{\ell}$ are defined by the Ohm law equations:
\begin{eqnarray}
\label{eq:Power_equations_appendix}
\left\{
\begin{array}{lll}
{p}^{\ell}_{or} = {|v^{\ell}_{or}|} {|v^{\ell}_{ex}|} (g^{\ell}\cdot \cos ({\theta^{\ell}_{or}} - {\theta^{\ell}_{ex}}) + b^{\ell} \sin ({\theta^{\ell}_{or}} - {\theta^{\ell}_{ex}}))\vspace{0.2cm} \\
{q}^{\ell}_{or} = {|v^{\ell}_{or}|} {|v^{\ell}_{ex}|} (g^{\ell}\cdot \sin ({\theta^{\ell}_{or}} - {\theta^{\ell}_{ex}}) - b^{\ell} \cos ({\theta^{\ell}_{or}} - {\theta^{\ell}_{ex}}))
\end{array}
\right.
\end{eqnarray}

The Kirchhoff energy conservation laws  states that at a node $k$ input power equals to output power, meaning that the injected power shall be balanced by power flows.

\begin{eqnarray}
\label{eq:Kirchhoff_energy_conservation}
{
\left\{
\begin{array}{lll}
0 = -&{p_k} + \sum_{s=1}^K {p}^{\ell}_{k} & \text{Active power}\vspace{0.2cm} \\
0 = &{q_k} + \sum_{s=1}^K {q}^{\ell}_{k}& \text{Reactive power}
\end{array}
\right.}
\end{eqnarray}

Finally, the global conservation law states the following:
\begin{equation}
\label{eq:global_conservation_law}
    Prod - Load =  (\sum_{\ell=1}^L (\hat{p}^{\ell}_{ex} + \hat{p}^{\ell}_{or}))
\end{equation}

This problem is non linear and non convex. To estimate these variables, a power flow solver such as the open-source package LightSim2grid \footnote{https://github.com/BDonnot/lightsim2grid} can be used. From these variables that are given by the simulators, it is then possible to derive other types of variables, such as the active and reactive power or the current flowing on all of the powerlines. 

\newpage
\section{Scenario design and data generation using LIPS}
\label{app:generation_example}
In this section, we show a practical example of how to design a scenario and how to generate the corresponding data using LIPS framework.

In order to promote the reproducibility and to facilitate the scenario design and data generation, the \textit{Data} module of LIPS framework (see Section \ref{par:lips} for more information) implements and provides a set of functionalities and integrates a physics solver. The scenarios could be easily designed through a configuration file accessible to the users which is read by a new agent allowing the implementation of various types of actions (e.g., multiple power line disconnections or topological changes) on the grid. The scenario could be specialized for each dataset, and the corresponding data are generated using the integrated physics solver.

The developed agent called \texttt{XDepthAgent} is used as a facilitator for the generation of datasets. It is integrated in data module of LIPS framework and under power grid utilities. It takes a grid2op environment, a list of required topological actions, the reference actions parameters, the scenario-specific parameters and offers multiple functionalities. For example, it allows to perform actions at different depths, combine various actions (operating on the bus bar or on lines) in the same scenario and also to ensure the legal combination and validity of the suggested actions. It allows also to combine the reference and scenario-based parameters which is required for the design of the competition scenarios. 

To generate some data using LIPS framework, the first step consists in providing a configuration file, which is used and interpreted by different modules and functions of LIPS framework for generation of dataset based on the corresponding scenario. This configuration file has two sections:
\newpage
\begin{itemize}
    \item \texttt{DEFAULT} This section is used to configure the global configurations required to parameterize the environment and solver. As can be seen, we declare the environment that should be used, its specific parameters such as the maximum number of power lines whose status could be changes. We could furthermore configure the chronics from which the data distribution are sampled and that for each of training and evaluation datasets separately. Finally, we can also set the seed number for reproducibility.

\begin{lstlisting}[language=json,firstnumber=1, caption={Configuration file: DEFAULT Section.},captionpos=b]
[DEFAULT]
env_name = "l2rpn_neurips_2020_track1_small"
env_params = {
	"NO_OVERFLOW_DISCONNECTION": True,
	"MAX_LINE_STATUS_CHANGED": 999999,
	"MAX_SUB_CHANGED": 999999,
	"NB_TIMESTEP_COOLDOWN_LINE": 0,
	"NB_TIMESTEP_COOLDOWN_SUB": 0}
chronics = {
	"train": "^((?!(.*[3-4][0-9].*)).)*$",
	"val": ".*3[0-5].*",
	"test": ".*3[5-9].*",
	"test_ood": ".*4[0-9].*"
	}
samples_per_chronic = {
	"initial_chronics_id": 0,
	"train": 864,
	"val": 288,
	"test": 288,
	"test_ood": 288,
	}
benchmark_seeds = {
	"train_env_seed": 1,
	"val_env_seed": 2,
	"test_env_seed": 3,
	"test_ood_topo_env_seed": 4,
	"train_actor_seed": 5,
	"val_actor_seed": 6,
	"test_actor_seed": 7,
	"test_ood_topo_actor_seed": 8,
	}
\end{lstlisting}
    \newpage
    \item \texttt{BENCHMARK} In this section, we can configure the attributes that should be used as inputs and outputs of the models and also configure more closely the agent behavior that is used to integrate required changes in data generation.

\begin{lstlisting}[language=json,firstnumber=1, caption={Configuration file: BENCHMARK section.}, captionpos=b] 
[Benchmark_competition]
attr_x = ("prod_p", "prod_v", "load_p", "load_q")
attr_tau = ("line_status", "topo_vect")
attr_y = ("a_or", "a_ex", "p_or", "p_ex", "v_or", "v_ex", "theta_or", "theta_ex")
attr_physics = ("YBus", "SBus", "PV_nodes", "slack")

dataset_create_params = {
	# REFERENCE PARAMS
	"reference_args" : {
		#"lines_to_disc": [3],
		"topo_actions": [
				{'set_bus':{'substations_id':[(1,(2,2,1,1,2,2))]}},#sub1
				{'set_bus':{'substations_id':[(16,(1,1,1,1,1,2,2,1,1,2,2,1,1,1,1,1,1))]}},#sub16
				{'set_bus':{'substations_id':[(16,(1,2,2,1,1,2,2,1,1,1,1,1,1,1,1,1,1))]}},#sub16
            {'set_bus':{'substations_id':[(28,(2,1,2,1,1))]}}#sub28
			],
		"prob_depth": (.5, .5),
		"prob_type": (1., 0.),
		"prob_do_nothing": .1,
		"max_disc": 0},
	# SCENARIO PARAMS
	"train": {
		# SCENARIO TOPOLOGY : disconnect or not one line at each tim step
		"prob_depth": (1.,), # sample from depth 1
		"prob_type": (0., 1.), # sample only from line disconnection
		"prob_do_nothing": 0.3,  # probability of do nothing
		"max_disc": 1}, # authorize at most 1 disconnection
	"test":{
		# SCENARIO TOPOLOGY: disconnect one line at each time step
		"prob_depth": (1.,), # sample from depth 1
		"prob_type": (0., 1.), # sample only from line disconnection
		"prob_do_nothing": 0.,  # No do nothing
		"max_disc": 1}, # authorize at most 1 disconnection
	"test_ood":{
		# SCENARIO TOPOLOGY: disconnect two lines at each time step
		"prob_depth": (0., 1.), # Sample only from depth 2
		"prob_type": (0., 1.), # sample only from line disconnection
		"prob_do_nothing": 0,  # No do nothing
		"max_disc": 2} # authorize at most 2 disconnection
	}
eval_dict = {
	"ML": ["MSE_avg", "MAE_avg", "MAPE_avg", "MAPE_90_avg", "MAPE_10_avg", "TIME_INF"],
	"Physics": ["CURRENT_POS", "VOLTAGE_POS", "LOSS_POS", "DISC_LINES", "CHECK_LOSS", 
               "CHECK_GC", "CHECK_LC", "CHECK_JOULE_LAW"],
	"IndRed": ["TIME_INF"],
	"OOD": ["MSE_avg", "MAE_avg", "MAPE_avg", "MAPE_90_avg", "MAPE_10_avg", "TIME_INF"]}
eval_params = {
	"inf_batch_size": 59000,
	"EL_tolerance": 0.04,
	"GC_tolerance": 1e-3,
	"LC_tolerance": 1e-2,
	"JOULE_tolerance": 1e-2
	}
\end{lstlisting}
\end{itemize}

\newpage
\paragraph{\textbf{Data Generation}}
Once the set of configurations are set by the user for a specific scenario, the corresponding data could be easily generated using LIPS framework. The following script shows the data generation through LIPS. It consists of instantiating a \textsc{Benchmark} class from LIPS package by indicating the path to the configuration file alongside the section name to be used. Next, the a call to the \textsc{generate} function of this class allows the generation of datasets by indicating the required number of observations.

\begin{python}[caption={A script from LIPS to generate datasets for specific scenarios indicated in the configuration file located at \texttt{CONFIG\_PATH}}]
from lips.benchmark.powergridBenchmark import PowerGridBenchmark

benchmark = PowerGridBenchmark(benchmark_path=DATA_PATH, # path to store data
                                benchmark_name="Benchmark_competition", # A section in configuration file
                                load_data_set=False, 
                                config_path=CONFIG_PATH, # path to the configuration file
                                log_path=LOG_PATH) # path to the logs

benchmark.generate(nb_sample_train=int(3e5),
                    nb_sample_val=int(1e5),
                    nb_sample_test=int(1e5),
                    nb_sample_test_ood_topo=int(2e5),
                    do_store_physics=True,
                    store_as_sparse=True
                   )
\end{python}

\newpage
\section{Score formulation}
\label{app:score_formulation}
The \textit{ML-related} and \textit{Physical compliance} criteria are computed separately on test and OOD test datasets. The \textit{speed-up} is computed only on test dataset as the inference time may remains same throughout the different dataset configurations. 

Hence, the global score is calculated based on a linear combination  of the introduced evaluation criteria categories on these datasets:
\begin{eqnarray}
\label{eq:score_appendix}
    \text{Score} = \alpha_{test} \times \text{Score}_{{test}} + \alpha_{ood} \times \text{Score}_{{OOD}} + \alpha_{speed-up} \times \text{Score}_{speed-up},
\end{eqnarray}
\noindent where $\alpha_{test}$, $\alpha_{ood}$, and $\alpha_{speed-up}$ are the coefficients to calibrate the relative importance related to metrics computed on test, OOD test dataset and speed-up with respect to the physical solver baseline. The procedure on how to compute each of these three sub-scores is explained with more ample details in the following.

\begin{enumerate}[leftmargin=15pt]
    \item Score$_{test}$: This sub-score is calculated based on a linear combination of 2 categories, namely: \textit{ML-related} and \textit{Physics compliance}.
    \begin{eqnarray}
    \label{eq:test_score}
        \text{Score}_{{test}} = \alpha_{ML} \times \text{Score}_{{ML}} + \alpha_{Physics} \times \text{Score}_{{Physics}}
    \end{eqnarray}
    where $\alpha_{ML}$ and $\alpha_{Physics}$ are the coefficients to calibrate the relative importance of ML-related and Physics compliance categories respectively for test dataset.
    
    \begin{enumerate}[i)]
        \item Score$_{ML}$: 
            For each quantity of interest, the ML-related sub-score is calculated based on two thresholds that are calibrated to indicate if the metric evaluated on the given quantity gives unacceptable/acceptable/great result. It corresponds to a score of 0 point / 1 point / 2 points, respectively. Within the sub-cateogry, Let :
            \begin{itemize}
                \item \textcolor{red}{$N_r$}: the number of unacceptable results overall (number of red circles);
                \item \textcolor{orange}{$N_o$}: the number of acceptable results overall (number of orange circles);
                \item \textcolor{DarkGreen}{$N_g$}: the number of great results overall (number of green circles).
            \end{itemize}
            Let also $N$ to be given by $N = \textcolor{red}{N_r} + \textcolor{orange}{N_o} + \textcolor{DarkGreen}{N_g}$. The score expression is given by:            
            \begin{eqnarray}
                \text{Score}_{{ML}} = \frac{1}{2N} (2 \times \textcolor{DarkGreen}{N_g} + 1 \times \textcolor{orange}{N_o} + 0 \times \textcolor{red}{N_r})
            \end{eqnarray}            
            A perfect score is obtained if all the given quantities provide great results. Indeed, we would have $N = \textcolor{DarkGreen}{N_g}$ and $\textcolor{red}{N_r} = \textcolor{orange}{N_o} = 0$ which implies $Score_{ML} = 1$.
        \item Score$_{Physics}$: For Physics compliance score $Score_{Physics}$, the score is also calibrated based on 2 thresholds and gives 0/1/2 points, similarly to $Score_{ML}$, depending on the result provided by various considered metrics mentioned earlier.
    \end{enumerate}
    
    \item Score$_{OOD}$: Exactly the same procedure as above for computation of $Score_{test}$ is used to compute the score on the out-of-distribution dataset using two evaluation categories which are : \textit{ML-related} and \textit{Physics compliance}. Hence, the $Score_{OOD}$ is obtained by:

    \begin{eqnarray}
    \label{eq:score_ood}
        \text{Score}_{{OOD}} = \alpha_{ML} \times \text{Score}_{{ML}} + \alpha_{Physics} \times \text{Score}_{{Physics}}
    \end{eqnarray}
    where $\alpha_{ML}$ and $\alpha_{Physics}$ are the coefficients to calibrate the relative importance of ML-related and Physics compliance categories respectively for out-of-distribution test dataset.
    
    \item Score$_{speed-up}$: For the speed-up criteria, we calibrate the score using a Weibull ``stretched exponential" function as follows:
    \begin{eqnarray}
    \label{eq:score_speedup}
        Score_{Speed} = \min \left(1. - exp((-\frac{x}{a})^b) , 1\right)
    \end{eqnarray}
    with $a = c\times(-\ln 0.9)^{-1/b}\quad \text{and}\quad b=1.7, c=5$. The $Score_{Speed}$ curve has the following shape:
    \begin{figure}[H]
        \centering
        \includegraphics[width=0.6\linewidth]{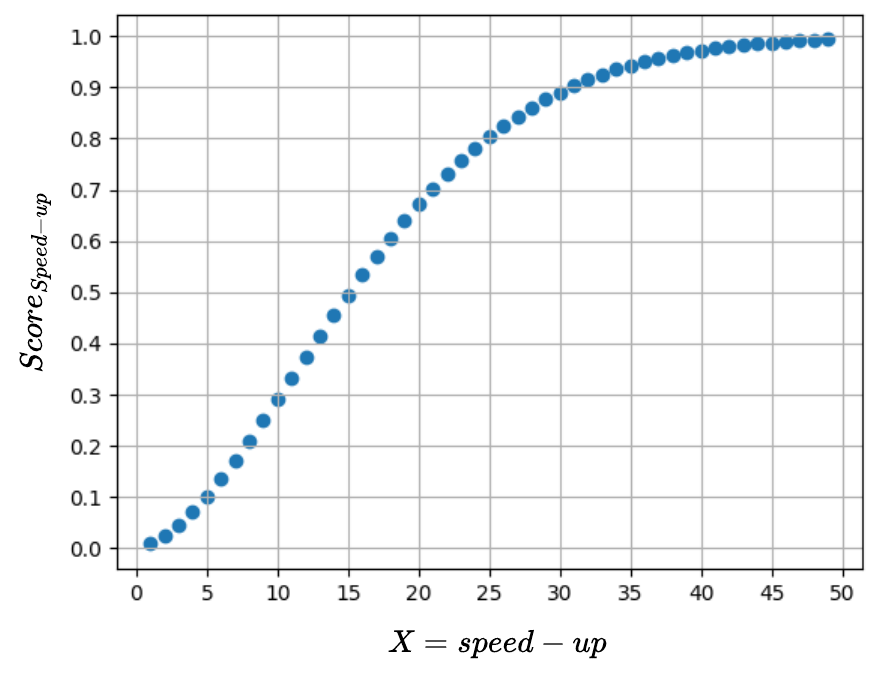}
        \caption{Speed-up score curve.}
        \label{fig:speedup_score}
    \end{figure}
    The $Speed-up$ is given by the ratio between the time required for a classical solver  to compute the power flow $time_{ClassicalSolver}$ and the inference time of an augmented simulator $time_{Inference}$:
    \begin{eqnarray}
        Speed-up = \frac{time_{ClassicalSolver}}{time_{Inference}}
    \end{eqnarray}
    The computed \( speed\text{-}up \) value is input into the Weibull function to determine the corresponding speed-up score, as shown on the vertical axis in Figure~\ref{fig:speedup_score}. This behavior promotes early acceleration of augmented simulators up to an acceleration factor of approximately 30, beyond which further increases have a minimal impact on the speed-up score.
\end{enumerate}

Various practical examples on how to compute the score for different baseline methods are shown in Appendix \ref{app:score_example}.

\newpage
\section{Thresholds used for discretization of metrics}
\label{app:thresholds}
We have opted for discretization of the evaluation scores for two reasons: 
\begin{itemize}
    \item To obtain a global score by aggregating over a set of heterogeneous set of evaluation criteria categories (described in the previous section);
    \item To obtain a synthetic visualization of the real valued metrics, which facilitates the reading of the benchmark table for a non-expert reader (see Table \ref{tab:Benchmark_evaluation_circle}. 
\end{itemize}

To discretize the evaluation metrics (described in Section \ref{par:lips}) into three modalities (green: great, orange: acceptable and red: not acceptable), we used two thresholds per evaluation metric by exploiting the power grid expert knowledge. All the evaluation metrics are considered as errors, where their minimization is preferable. The Tables \ref{tab:physics_threshold} and \ref{tab:ml_threshold} summarize these thresholds for physics and machine learning evaluation metrics correspondingly. 

\begin{table}[H]
    \centering
    \caption{Thresholds table for physics compliance metrics. The physical laws values are discretized into three intervals on the basis of theses thresholds. The values below the inferior threshold are considered as perfect. The values between the inferior and superior thresholds are considered as acceptable, and the values above the superior threshold are considered as not acceptable.}
    \resizebox{0.9\linewidth}{!}{
    \begin{tabular}{cc|ccccccccc}
         \toprule
         & \textbf{Physics compliances} && P1 & P2 & P3 & P4 & P5 & P6 & P7 & P8\\
         \midrule
         \multicolumn{1}{c|}{\multirow{2}{*}{\rotatebox{90}{Thresholds}}} & Inferior && 1\% & 1\% & 1\% & 1\% & 1\% & 5\% & 5\% & 1\%\\ \multicolumn{1}{c|}{} & \multicolumn{1}{c|}{}
         \\ \cline{2-11} \multicolumn{1}{c|}{} & \multicolumn{1}{c|}{} \\  \multicolumn{1}{c|}{} & Superior && 5\% & 5\% & 5\% & 5\% & 5\% & 10\% & 10\% & 5\%\\
         \bottomrule
    \end{tabular}}
    \label{tab:physics_threshold}
\end{table}

\begin{table}[H]
    \centering
    \caption{Thresholds table for Machine Learning (ML) related metrics. The ML metric values are discretized into three intervals on the basis of theses thresholds. The values below the inferior threshold are considered as perfect. The values between the inferior and superior thresholds are considered as acceptable, and the values above the superior threshold are considered as not acceptable.}
    \begin{tabular}{cc|ccccccc}
         \toprule
         & \textbf{ML-related} && $a_{or}$ & $a_{ex}$ & $a_{or}$ & $a_{ex}$ & $a_{or}$ & $a_{ex}$\\
         \midrule
         \multicolumn{1}{c|}{\multirow{2}{*}{\rotatebox{90}{Thresholds}}} & Inferior && 2\% & 2\% & 2\% & 2\% & 0.2 & 0.2\\ \multicolumn{1}{c|}{} & \multicolumn{1}{c|}{}
         \\ \cline{2-9} \multicolumn{1}{c|}{} & \multicolumn{1}{c|}{} \\  \multicolumn{1}{c|}{} & Superior && 5\% & 5\% & 5\% & 5\% & 0.5 & 0.5\\
         \bottomrule
    \end{tabular}
    \label{tab:ml_threshold}
\end{table}
\newpage
\section{Practical example to compute the score}
\label{app:score_example}
Using the notation introduced in the previous subsection, let us consider the following configuration:
\begin{itemize}
    \item $\alpha_{\text{test}} = 0.3$  
    \item $\alpha_{\text{ood}} = 0.3$  
    \item $\alpha_{\text{speed-up}} = 0.4$  
    \item $\alpha_{ML} = 0.66$  
    \item $\alpha_{Physics} = 0.34$  
    \item $b=1.7, c=5$
\end{itemize}
In order to illustrate even further how the score computation works, we provide in Table 2 examples for the load flow prediction task.

\subsection{Score computation for Security Analysis}
As it is the most straightforward to compute, we start with the global score for the solution obtained with 'Grid2Op', the physical solver used to produce the data. It is the reference physical solver, which implies that the accuracy is perfect but the speed-up is lower than the expectation. For illustration purpose, we use the speedup obtained by security analysis which was $3.7$ faster than the Grid2op solver. Therefore, we obtain the following sub-scores:

\begin{itemize}
    \item $Score_{test} = 0.66 \times \left(\frac{2 \times 6}{2 \times 6}\right) + 0.34 \times \left(\frac{2 \times 8}{2 \times 8}\right) = 1$
    \item $Score_{ood} = 0.66 \times \left(\frac{2 \times 6}{2 \times 6}\right) + 0.34 \times \left(\frac{2 \times 8}{2 \times 8}\right) = 1$
    \item $Score_{speedup} = 1 - exp(-\frac{3.77}{18.78})^{1.7} = 0.06$
\end{itemize}
Then, by combining them, the global score is $Score_{PhysicsSolver} = 0.3 \times 1 + 0.3 \times 1 + 0.4 \times 0.06 = 0.625$, therefore 62.5\%.\\

\subsection{Score computation for LEAPNet}
The procedure is similar with LeapNet architecture. The associated subscores are:
\begin{itemize}
    \item $Score_{test} = 0.66 \times \left(\frac{2 \times 2 + 1 \times 2 + 0 \times 2}{2 \times 6}\right) + 0.34 \times \left(\frac{2 \times 2 + 1 \times 1 + 0 \times 5}{2 \times 8}\right) = 0.44$
    \item $Score_{ood} = 0.66 \times \left(\frac{2 \times 0 + 1 \times 4 + 0 \times 2}{2 \times 6}\right) + 0.34 \times \left(\frac{2 \times 2 + 1 \times 1 + 0 \times 5}{2 \times 8}\right) = 0.33$
    \item $Score_{speedup} = 1 - exp(-\frac{11.9}{18.78})^{1.7} = 0.36$
\end{itemize}
Then, by combining them, the global score is $Score_{LeapNet} = 0.3 \times 0.44 + 0.3 \times 0.33 + 0.4 \times 0.36 = 0.376$, therefore 37.6\%.

\bibliographystyle{elsarticle-num} 
\bibliography{references}

\end{document}